\documentclass[journal]{IEEEtran}

\usepackage{graphicx}
\usepackage{caption}
\usepackage{cite}
\usepackage{amsmath}
\usepackage{bm}
\usepackage{amssymb}
\usepackage{tikz}
\usepackage{color}
\usepackage{stfloats}
\begin{document}
\title{Learning Ultrasound Scanning Skills\\ from Human Demonstrations}
\author{Xutian Deng, Ziwei Lei, Yi Wang and Miao Li\IEEEauthorrefmark{2}
\thanks{{All the authors are with Wuhan University, Wuhan, Hubei, China. (email:
dengxutian@whu.edu.cn)}}
\thanks{\IEEEauthorrefmark{2} Miao Li is the corresponding author. (email: miao.li@whu.edu.cn, limiao712@gmail.com)}
}
\maketitle

\begin{abstract}
Recently, the robotic ultrasound system has become an emerging topic owing to the widespread use of medical ultrasound. However, it is still a challenging task to model and to transfer the ultrasound skill from an ultrasound physician. In this paper, we propose a learning-based framework to acquire ultrasound scanning skills from human demonstrations. First, the ultrasound scanning skills are encapsulated into a high-dimensional multi-modal model in terms of interactions among ultrasound images, the probe pose and the contact force. The parameters of the model are learned using the data collected from skilled sonographers' demonstrations. Second, a sampling-based strategy is proposed with the learned model to adjust the extracorporeal ultrasound scanning process to guide a newbie sonographer or a robot arm. Finally, the robustness of the proposed framework is validated with the experiments on real data from sonographers.           
\end{abstract}

\IEEEpeerreviewmaketitle

\section{Introduction}

Medical ultrasound imaging technology is widely adopt in clinical diagnosis due to its  incomparable merits including non-invasive, low-hazard, real-time imaging,  safe and low cost. Nowadays, ultrasound imaging can quickly detect diseases of different anatomical structures, including liver \cite{gerstenmaier2014ultrasound}, gallbladder \cite{konstantinidis2012gallbladder}, bile duct \cite{lahham2018utility}, spleen \cite{omar2016contrast}, pancreas \cite{larson2016ultrasound}, kidney \cite{correas2016ultrasound}, adrenal gland \cite{dietrich2010contrast}, bladder \cite{daurat2015diagnosis}, prostate \cite{mitterberger2010ultrasound}, thyroid \cite{haymart2019thyroid}, etc. In addition, during the global pandemic of COVID-19, medical ultrasound is also used for the diagnosis of infected persons by detecting pleural effusion \cite{buonsenso2020covid, soldati2020proposal}. However, the performance of ultrasound examination is highly depending on skills and experience of sonographers, which generally require a large amount of time and effort to acquire \cite{arger2005teaching, hertzberg2000physician}. Moreover, the intensive and repetitive ultrasound scanning process causes a heavy burden on sonographers' physical condition, further leading to the scarcity of ultrasound practitioners. 

To address these issues, many previous studies in robotics have attempt to use robots to help or even to replace the sonographers \cite{boctor2008three, priester2013robotic, chatelain20153d}. According to the level of the system autonomy, robotic ultrasound system can be categorized into three levels: tele-operated, semi-autonomous and full-autonomous. A tele-operated robotic ultrasound system usually contains two main parts: teacher site and student site \cite{seo2015development, mathiassen2016ultrasound, guan2017study}. The motion of the student robot is completely determined by the teacher, usually a trained sonographer through different kinds of interaction devices including 3D space mouse \cite{seo2015development}, inertial measurement unit (IMU) handle \cite{guan2017study, sandoval2020cobot}, haptic interface \cite{sandoval2020cobot}, etc. While for a semi-autonomous robotic ultrasound system, the motion of the student robot is partly determined by the teacher \cite{patlan2017robotic, mathur2019semi, victorova20193d}. 

For a full-autonomous robotic ultrasound system, the student robot is supposed to perform the whole process of local ultrasound scanning by itself \cite{virga2016automatic, kim2017development, huang2018robotic} and the teacher robot is only used for emergency or unexpected situations. Until today, only local full-autonomous robotic ultrasound systems have been reported \cite{hennersperger2016towards, ning2021autonomic}. These robotic ultrasound systems usually focus on scanning of some certain anatomical structures, such as abdomen \cite{hennersperger2016towards}, thyroid \cite{kim2017development} and vertebra \cite{ning2021autonomic}. Despite of these achievements, it is still a challenging task to represent and learn the ultrasound scanning skill due to its high dimensionality and rich modality. A comprehensive survey on robotic ultrasound is given in TABLE \ref{table:summary}. 

\begin{figure}[t]
\centering
\includegraphics[width=1\linewidth]{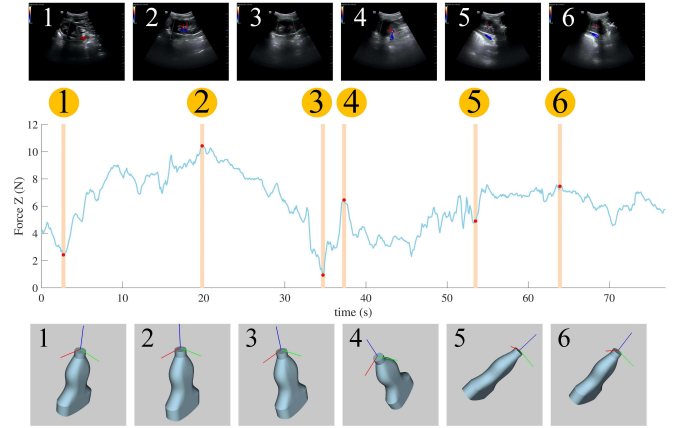}
\caption{The feedback information from three different modalities during a free-hand ultrasound scanning process. The first row represents ultrasound images. The second row represents the contact force in the z-axis between the probe and the skin, collected using a 6 dimensional force/torque sensor. The third row represents the probe pose, which is collected using an inertial measurement unit (IMU).}
\label{fig_1::intro}
\end{figure}

\begin{table*}[ht]
\setlength{\tabcolsep}{2mm}
\centering
\caption{A brief summary of robotic ultrasound. Initials: convolutional neural network (CNN), magnetic resonance imaging (MRI), support vector machine (SVM), reinforcement learning (RL).}
\begin{tabular}{|l|lllll|}
\cline{1-6}
Paper                           & Autonomy Degree & Specific Target & Modality                                            & Guidance                    & Publication Year   \\ 
\cline{1-6}
\cite{seo2015development}       & tele-operated   & no              & force, orientation, position                     & human                       & 2015   \\
\cite{mathiassen2016ultrasound} & tele-operated   & no              & force, orientation, position                     & human                       & 2016   \\
\cite{guan2017study}            & tele-operated   & no              & force, orientation, position                     & human                       & 2017   \\
\cite{sandoval2020cobot}        & tele-operated   & no              & force, orientation, position                     & human                       & 2020   \\
\cite{patlan2017robotic}        & semi-autonomous & no              & force, orientation, position, elastogram         & elastogram, human           & 2017   \\
\cite{mathur2019semi}           & semi-autonomous & no              & force, orientation, position, vision             & CNN, human                  & 2019   \\ 
\cite{victorova20193d}          & semi-autonomous & yes             & force, orientation, position                     & trajectory, human           & 2019   \\
\cite{kim2020robot}             & semi-autonomous & yes             & force, orientation, position, image              & CNN, human                  & 2020   \\
\cite{virga2016automatic}       & full-autonomous & yes             & force, orientation, position, vision, image, MRI & vision, MRI, confidence map & 2016   \\
\cite{kim2017development}       & full-autonomous & yes             & force, orientation, position, image              & SVM                         & 2017   \\
\cite{huang2018robotic}         & full-autonomous & no              & force, orientation, position, vision             & vision                      & 2018   \\
\cite{hennersperger2016towards} & full-autonomous & yes             & force, orientation, position, vision, MRI        & vision, MRI                 & 2016   \\
\cite{ning2021autonomic}        & full-autonomous & yes             & force, position, vision                          & RL      & 2021   \\
\cline{1-6}
\end{tabular}
\label{table:summary}
\end{table*}

\begin{figure*}[bp]
\centering
\includegraphics[width=1\linewidth]{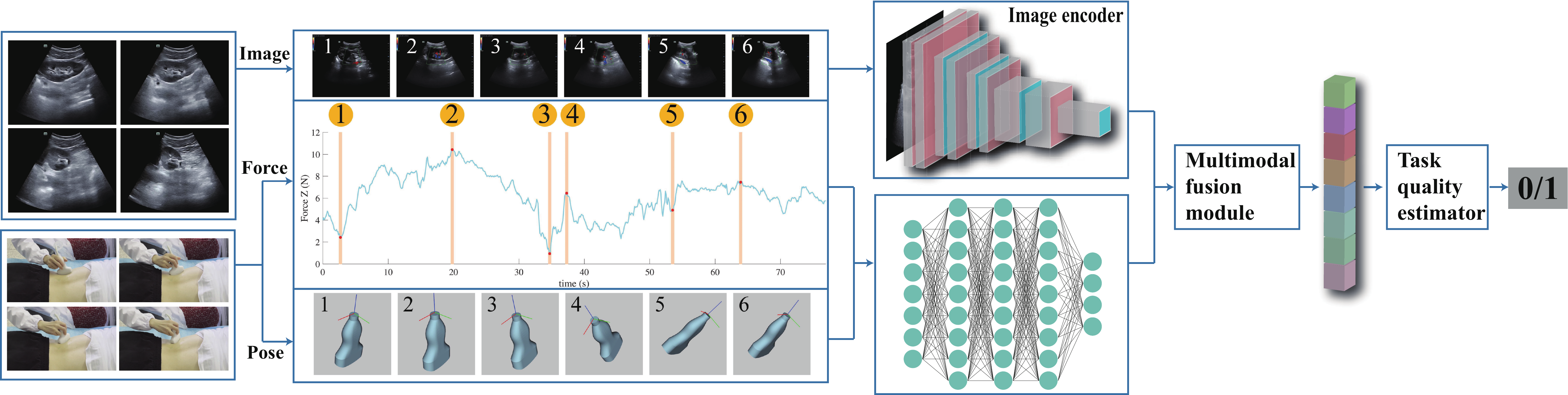}
\caption{The multi-modal task learning architecture with human annotations. The network takes data from three different sensors as input: the ultrasound images, force/torque (F/T) and the pose information. The data for the task learning is acquired through human demonstrations, where the ultrasound quality is evaluated by sonographers. With the trained network, the multi-modal task can be represented as a high-dimensional vector.}
\label{fig::task_rep}
\end{figure*}

In this paper, we proposed a learning-based approach to represent and learn ultrasound skills from sonographers' demonstrations, in order to guide the scanning process. During the learning process, the ultrasound images together with the relevant scanning variables (the probe pose and the contact force) are recorded and encapsulated into a high-dimensional model. Then, we leverage the power of deep learning to implicitly capture the relation between the quality of ultrasound images and scanning skills. During the execution stage, the learned model is used to evaluate the current quality of the ultrasound image. In order to obtain a high quality ultrasound image, a sampling-based approach is used to adjust the probe motion.

The main contribution of this paper is twofold: 1. A multi-modal model of ultrasound scanning skills is proposed and learned from human demonstrations, which takes ultrasound images, the probe pose and the contact force into account. 2. Based on the learned model, a sampling-based strategy is proposed to adjust the ultrasound scanning process, in order to obtain a high quality ultrasound image. Note that the goal of this paper is to offer a learning-based framework to understand and acquire the ultrasound skills from human demonstrations. However, it is obvious that the learned model can be ported into a robot system as well, which is our work for the next step.

This paper is organized as follows. Section II presents related work in the field of ultrasound images and ultrasound scanning guidance. Section III provides the methodology of our model, including the learning process of task representation, the data acquisition process through human demonstrations and the strategy for the scanning guidance during real time execution. Section IV describes the detailed experimental validation, with a final discussion and conclusion in Section V.

\begin{figure*}[ht]
\centering
\includegraphics[width=1\linewidth]{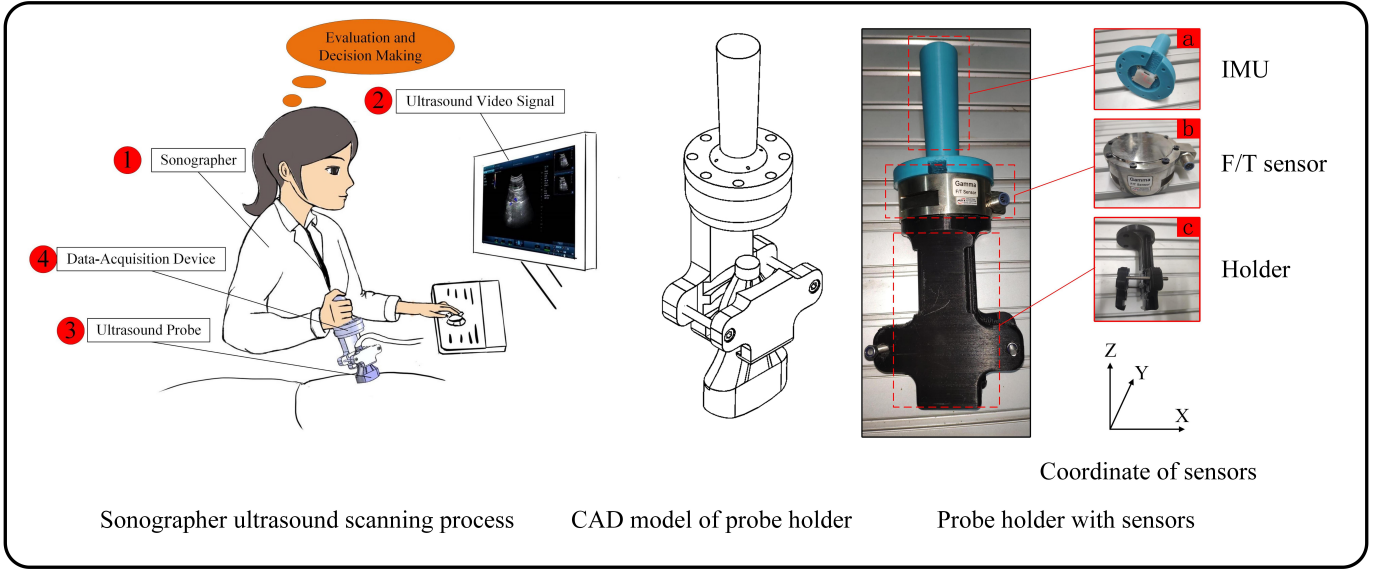}
\caption{The ultrasound scanning data collected from human demonstrations. The sonographer is performing an ultrasound scanning with a specifically designed probe holder. The sensory feedback during the scanning process is recorded, including the ultrasound images from an ultrasound machine, the contact force and torque from a 6D F/T sensor, and the probe pose from an IMU sensor.}
\label{fig::human_demo}
\end{figure*}

\section{Related Work}

\subsection{Ultrasound Images Evaluation}

The goal of the ultrasound images evaluation is to understand images in terms of classification \cite{hijab2019breast}, segmenting \cite{ghose2013supervised}, recognition \cite{wang2019automatic}, etc. With the rise of deep learning, many studies have attempt to process the ultrasound images with the help of neural networks.

Liu \emph{et al.} have summarized the extensive research results on ultrasound images processing with different network structures including convolution neural network (CNN), recurrent neural network (RNN), auto-encoder network (AE), restricted Boltzmann’s machine (RBM), deep belief network (DBN), etc \cite{liu2019deep}. From the perspective of applications, Sridar \emph{et al.} have employed the CNN for the main plane classification in fetal ultrasound images, considering both local and global features of the ultrasound images \cite{sridar2019decision}. In order to judge the severity of patients, Roy \emph{el al.} have collected ultrasound images of the COVID-19 patient's lesions to train a spatial transformer network \cite{roy2020deep}. Deep learning is also adopted in the task of segmenting thyroid nodules from the real-time ultrasound images \cite{ouahabi2021deep}.
While deep learning provides a superior framework to understand the ultrasound images, it generally requires a large number of expert-labeled data, which can be difficult and expensive to collect.    

Confidence map provides an alternative method in ultrasound images processing \cite{karamalis2012ultrasound}. The confidence map is obtained through pixel-wise confidence estimation using a random walk. Chatelain \emph{et al.} have devised a control law based on the ultrasound confidence map \cite{chatelain2015optimization, chatelain2016confidence}, with the goal to adjust the in-plane rotation and motion of the probe.  Confidence map is also employed to automatically determine proper parameters for the ultrasound scanning \cite{virga2016automatic}.  
Furthermore, the advantages of the confidence maps have been demonstrated by combining with position control and force control to preform automatic position and pressure maintenance \cite{chatelain2017confidence}. However, confidence map is proposed with the hand-coded rules, which can not be directly used to guild the scanning motion.

\subsection{Learning of the Ultrasound Scanning Skills}
While the goal of the ultrasound images processing is to understand images, learning of the ultrasound scanning skills aims to obtain high-quality ultrasound images through the adjustment of the scanning operation. Droste \emph{et al.} have used a clamping device with IMU to obtain the relation between the probe pose and the ultrasound images during ultrasound examination \cite{droste2020automatic}. Li \emph{et al.} have built a simulation environment based on 3D ultrasound data acquired by robot arm mounted with a ultrasound probe \cite{li2021autonomous}. However, they didn't explicitly learn the ultrasound scanning skills. Instead, a reinforcement learning framework is adopted to optimize the confidence map of ultrasound images, by adapting the movement of the ultrasound probe. All of the above-mentioned work only take the pose and the position of the probe as input, while in this paper the contact force between the probe and humans is also encoded, which is considered as a crucial factor during ultrasound scanning process \cite{jiang2020automatic}.

For the learning of force-relevant skills, a great variety of previous studies in robotic manipulation focused on learning the relation between force information and other task-related variables, such as the position and velocity \cite{gao2019learning}, the surface electromyography \cite{zeng2020simultaneously}, the task states and constraints \cite{holladay2021planning}, and the desired impedance \cite{li2018learning,li2014learning,li2014adaptation}. A multi-modal representation method for contact-rich tasks has been proposed in \cite{lee2019making} to encode the concurrent feedback information from vision and touch. The method was learned through self-supervision, which can be further exploited to improve the sampling efficiency and the task success rate. To the best of our knowledge, for a multi-modal manipulation task including feedback information from ultrasound, force and motion, this is the first work to learn the task representation and the corresponding manipulation skills from human demonstrations.

\section{Problem Statement and Method}
Our goal is to learn the free-hand ultrasound scanning skills from human demonstrations. We want to evaluate the multi-modal task quality of combining multiple sensory information including ultrasound images, the probe pose and the contact force, with the goal to extract skills from the task representation and even to transfer skills across tasks. We formulate the multisensory data by a neural network, where the parameters are trained by the data supervised through human ultrasound experts. In this section, we will discuss the learning process of the task representation, the data collection procedure and the online ultrasound scanning guidance respectively.

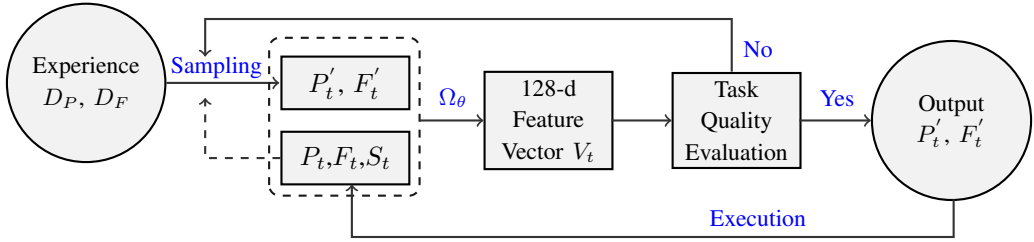
\begin{figure*}[ht]
\centering
\begin{tikzpicture}[%
block/.style = {draw, thick,fill=black!30, fill opacity=1,align=center, anchor=west,
            minimum height=0.7cm, inner sep=0.1cm},
ball/.style = {circle, draw,thick, fill opacity=1,align=center, anchor=west, inner sep=0}]

\node[ball,anchor=west,text width=2cm,fill=black!5] (data) at (0,0) {\small{Experience}\\$D_P$, $D_F$};
\draw[->,thick,draw=black!80] (data.east) to +(1.5,0);
\node at (2.8,0.2){\small \textcolor{blue}{Sampling}};
\node[block,anchor=west,text width=1.5cm,fill=black!5] (action)  at (3.65,0) {$P^{'}_t$, $F^{'}_t$ };
\node[block,anchor=west,text width=1.5cm,fill=black!5] (image)  at (3.65,-1) {$P_t$,$F_t$,$S_t$ };
\draw[rounded corners, draw=black!90,thick, dashed](3.5,-1.5) rectangle (5.5,0.6);

\node[block,anchor=west,text width=1.5cm,fill=black!5] (execution)  at (6.35,-0.5) {\small{128-d Feature Vector $V_t$}};
\node at (5.95,-0.2){\small \textcolor{blue}{$\Omega_\theta$}};
\node[block,anchor=west,text width=1.5cm,fill=black!5] (execution)  at (8.85,-0.5) {\small{Task Quality Evaluation} };
\draw[->,thick,draw=black!80] (5.5,-0.5) to +(0.85,0);
\draw[->,thick,draw=black!80] (8.05,-0.5) to +(0.8,0);
\draw[->,thick,draw=black!80] (10.55,-0.5) to +(0.95,0);
\node at (11.05,-0.2){\small \textcolor{blue}{\small{Yes}}};
\node[ball,anchor=west,text width=2cm,fill=black!5] (store) at (11.5,-0.5) {\small{Output}\\$P^{'}_t$, $F^{'}_t$};
\node at (10.0,0.45){\small \textcolor{blue}{\small{No}}};
\draw[->,thick,draw=black!80] (9.65,0.15) to +(0,0.7) to +(-7.0,0.7) to +(-7.0,0.2);
\node at (10.0,-1.8){\small \textcolor{blue}{\small{Execution}}};
\draw[->,thick,draw=black!80] (12.6,-1.55) to +(0,-0.5) to +(-8, -0.5) to +(-8, 0.2);

\draw[->,thick,draw=black!80,dashed] (3.65,-1.0) to +(-1,0) to +(-1,0.8);
\end{tikzpicture}

\caption{Our strategy for scanning guidance takes the current pose $P_t$, the contact force $F_t$ and the ultrasound image $S_t$ as input, and outputs the next desired pose $P^{'}_t$ and contact force $F^{'}_t$. For sampling, we impose a bound between $P^{'}_t$, $F^{'}_t$ and $P_t$, $F_t$, which prevents the next state from moving too far away from the current state. For execution, the desired pose $P^{'}_t$ and contact force $F^{'}_t$ is used as a goal for the human ultrasound scanning guidance.}
\label{fig::skill_sampling}
\end{figure*}

\begin{figure}[bp]
\centering
\includegraphics[width=1.0\linewidth]{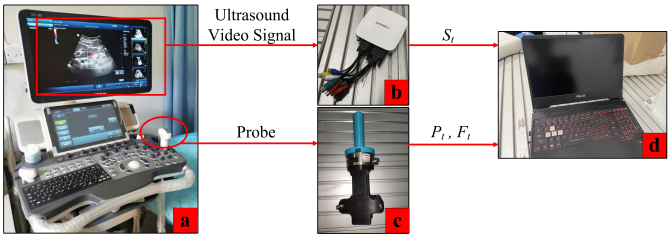}
\caption{Experiments setup. (a) The ultrasound machine - Mindray DC-70. (b) The video capture device - MAGEWELL USB Capture AIO. (c) Data-acquisition probe holder. (d) The computer for data collection with Intel i5 CPU and Nvidia GTX 1650 GPU, Ubuntu16.04 LTS.}
\label{fig::exp_setup}
\end{figure}

\begin{figure}[bp]
\centering
\begin{minipage}[t]{0.24\linewidth}
\centering
\includegraphics[width=1\linewidth]{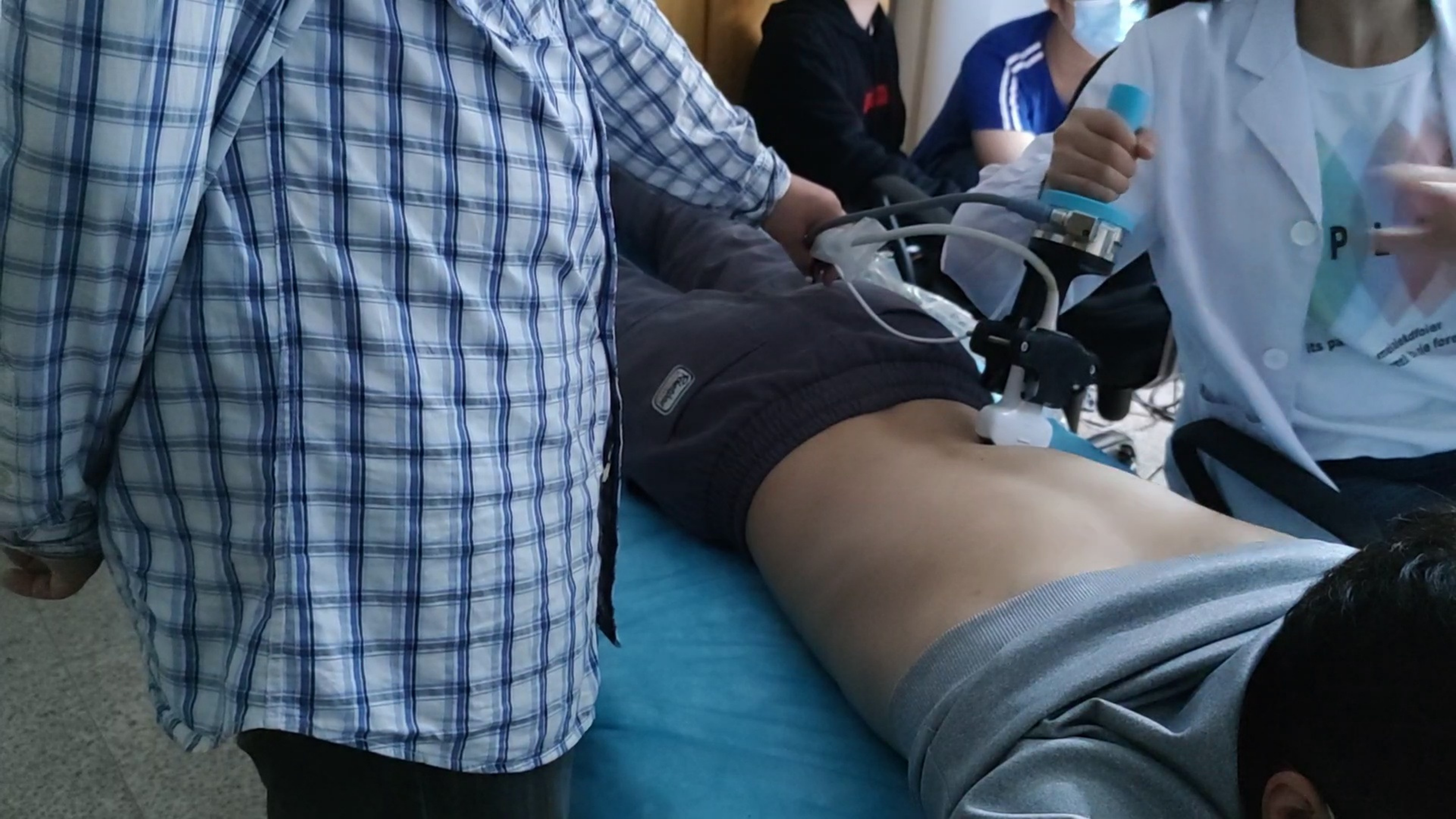}
\caption*{(a)}
\end{minipage}
\begin{minipage}[t]{0.24\linewidth}
\centering
\includegraphics[width=1\linewidth]{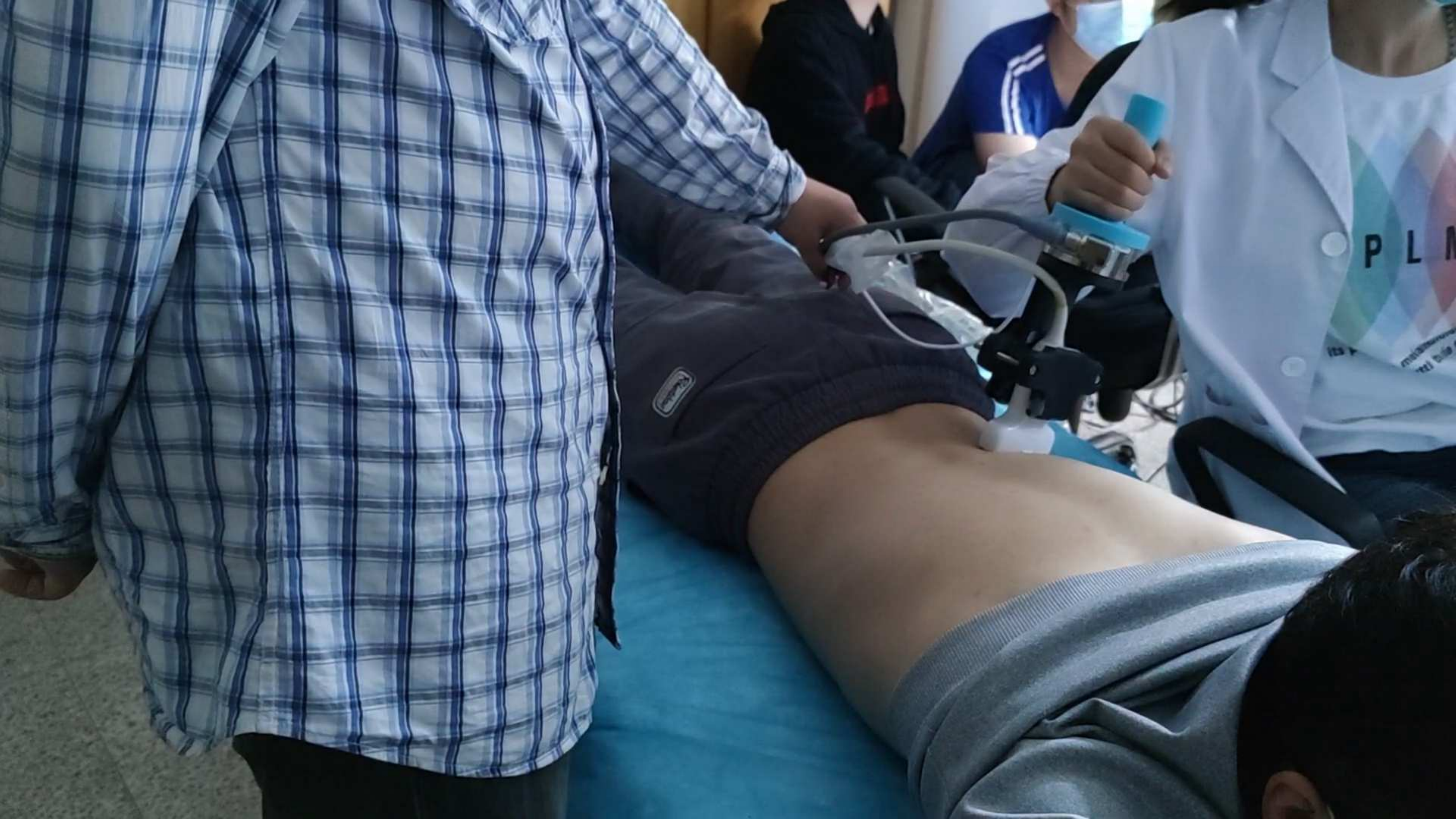}
\caption*{(b)}
\end{minipage}
\begin{minipage}[t]{0.24\linewidth}
\centering
\includegraphics[width=1\linewidth]{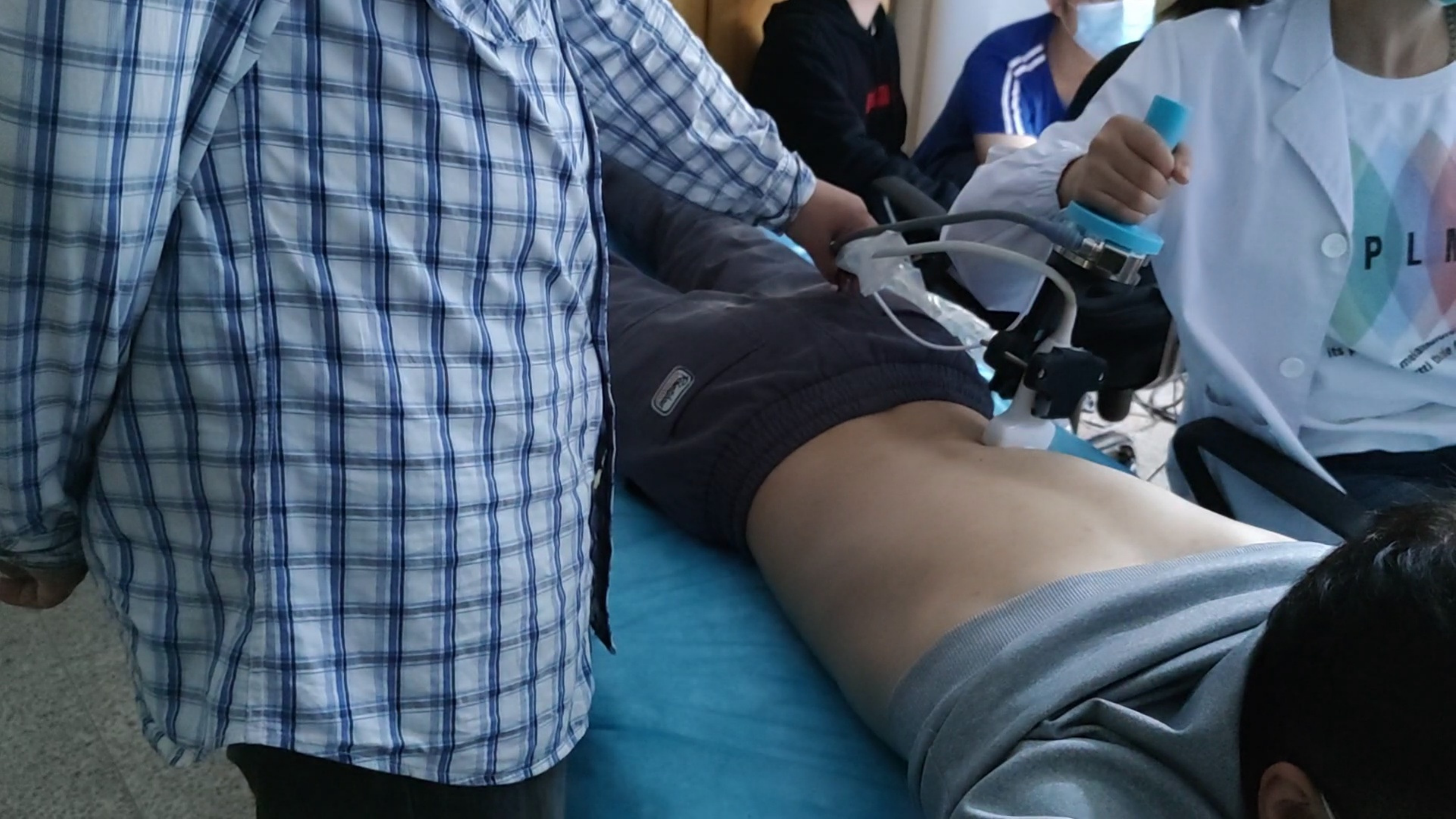}
\caption*{(c)}
\end{minipage}
\begin{minipage}[t]{0.24\linewidth}
\centering
\includegraphics[width=1\linewidth]{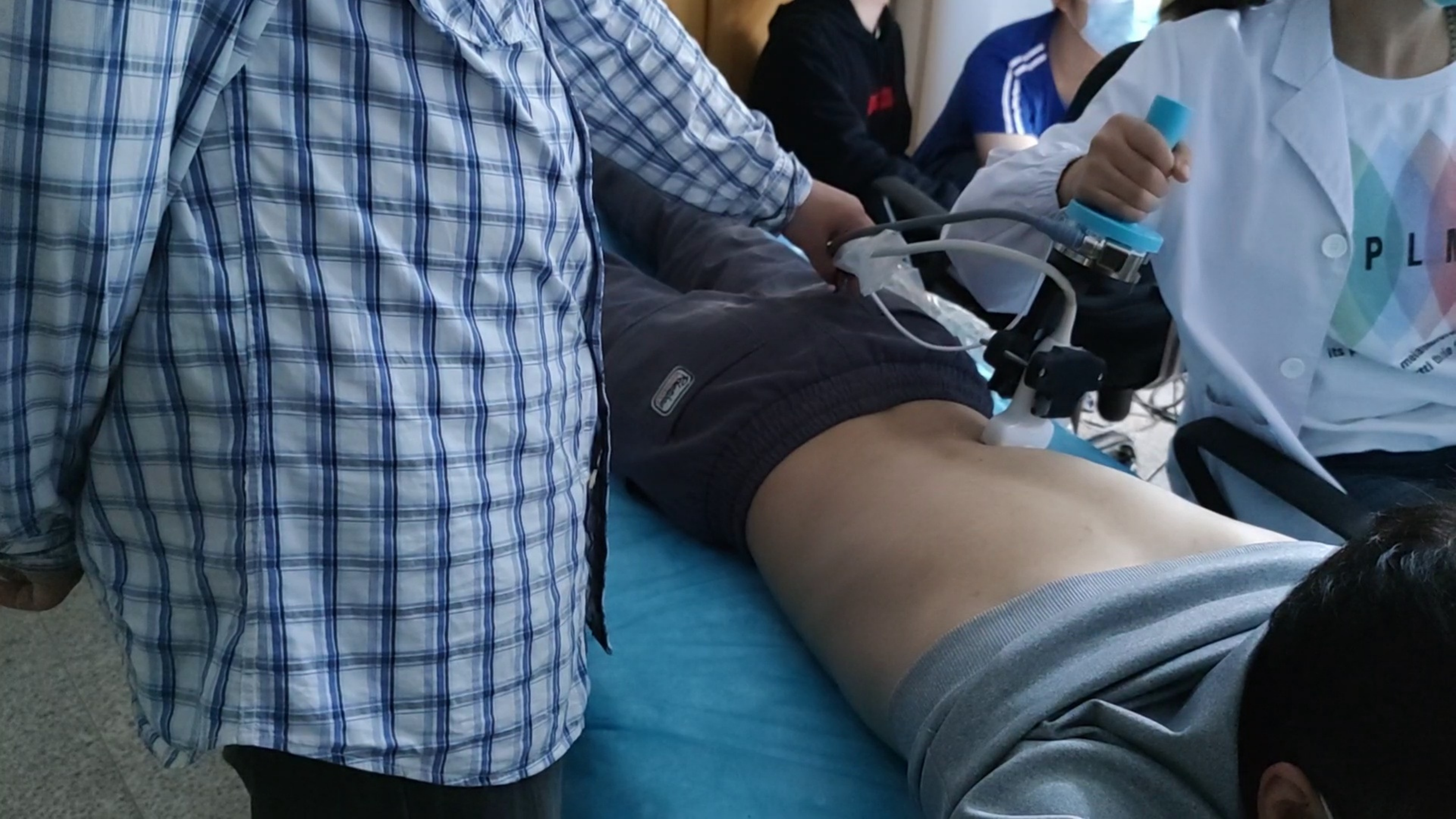}
\caption*{(d)}
\end{minipage}
\begin{minipage}[t]{0.24\linewidth}
\centering
\includegraphics[width=1\linewidth]{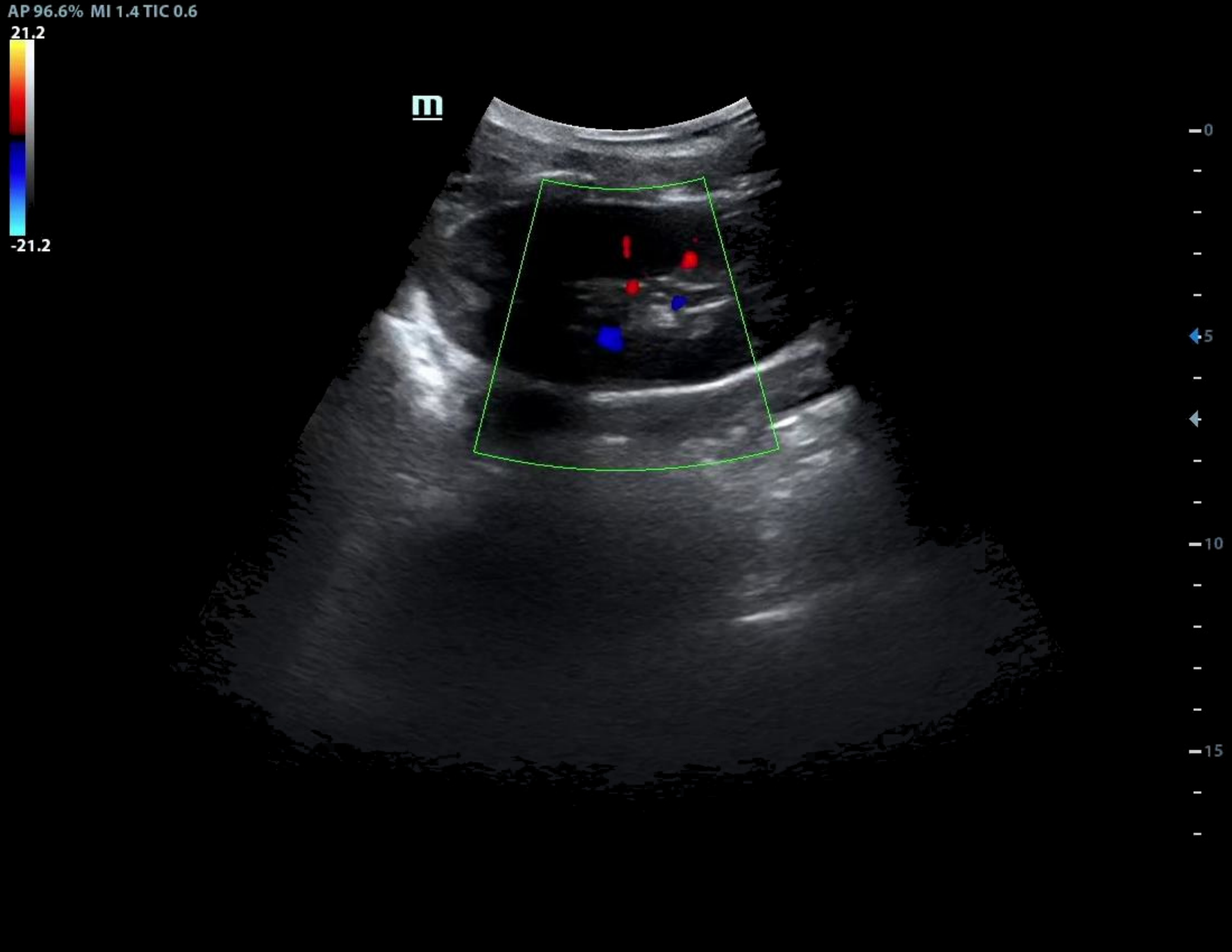}
\caption*{(e)}
\end{minipage}
\begin{minipage}[t]{0.24\linewidth}
\centering
\includegraphics[width=1\linewidth]{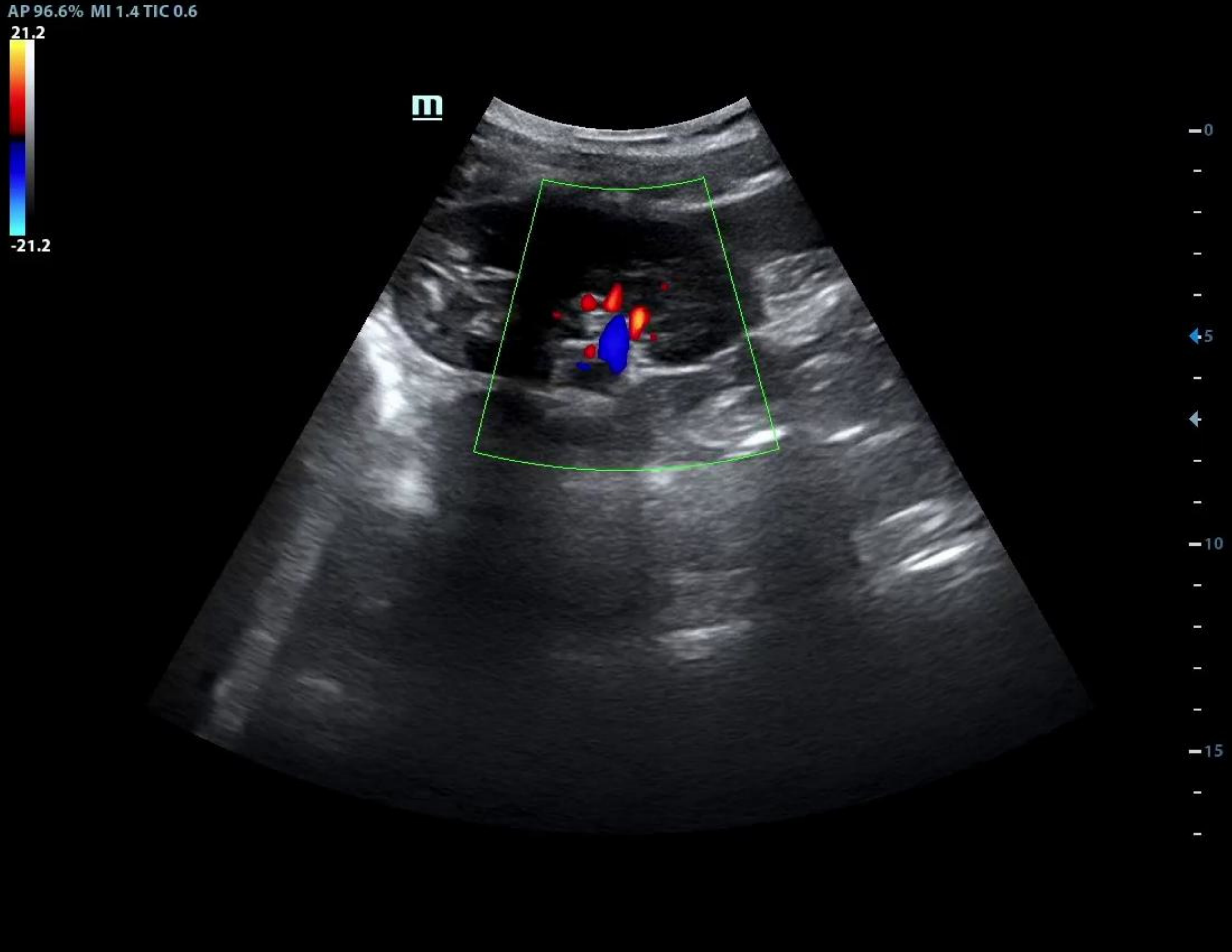}
\caption*{(f)}
\end{minipage}
\begin{minipage}[t]{0.24\linewidth}
\centering
\includegraphics[width=1\linewidth]{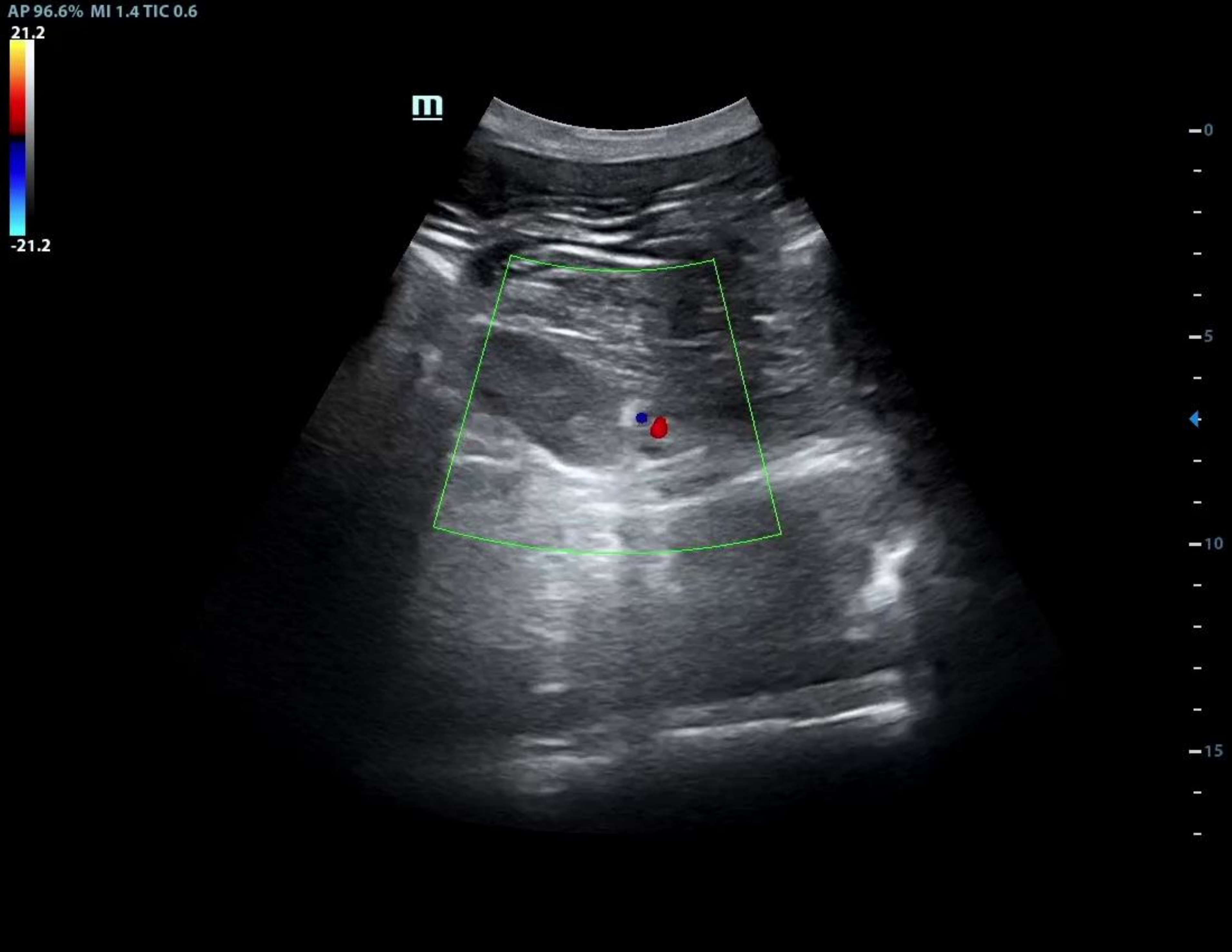}
\caption*{(g)}
\end{minipage}
\begin{minipage}[t]{0.24\linewidth}
\centering
\includegraphics[width=1\linewidth]{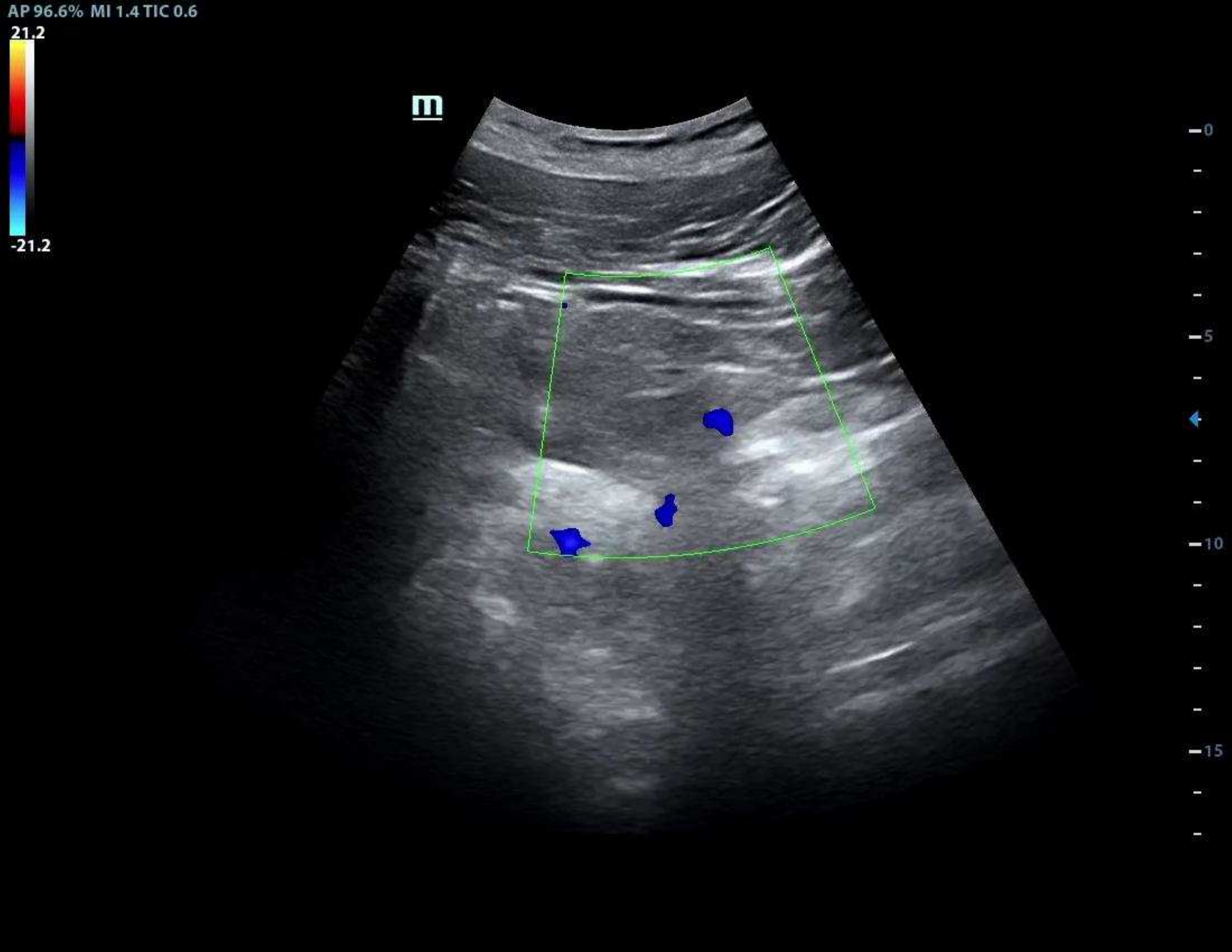}
\caption*{(h)}
\end{minipage}
\caption{The snapshots of human ultrasound scanning demonstrations and samples of the obtained ultrasound images. Here the images (e) and (f) are labeled as good quality while (g) and (h) are labeled as bad quality.}
\label{fig::human_scan}
\end{figure}

\begin{figure*}[ht]
\centering
\begin{minipage}[t]{0.30\linewidth}
\centering
\includegraphics[width=1.1\linewidth]{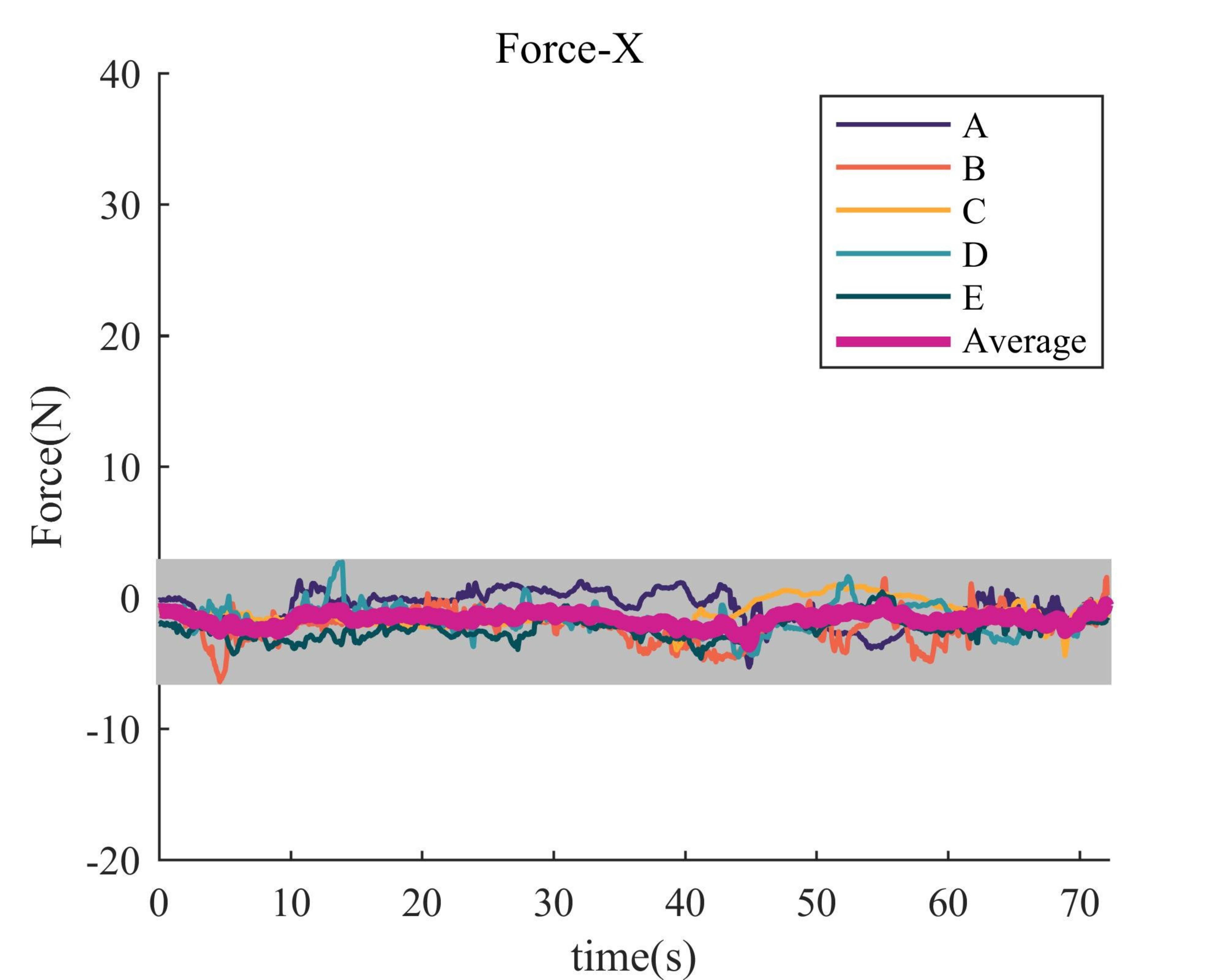}
\caption*{(a)}
\end{minipage}
\begin{minipage}[t]{0.30\linewidth}
\centering
\includegraphics[width=1.1\linewidth]{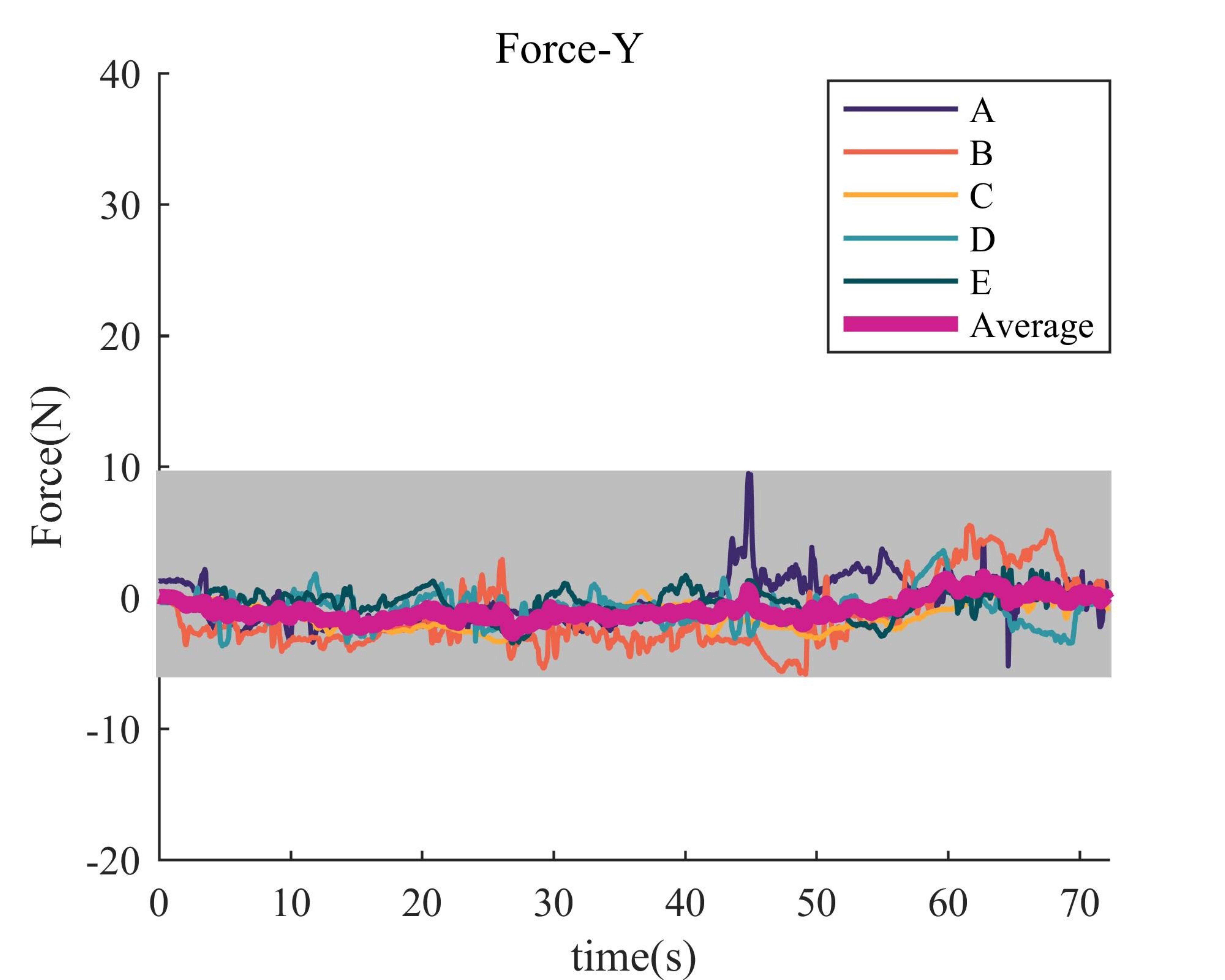}
\caption*{(b)}
\end{minipage}
\begin{minipage}[t]{0.30\linewidth}
\centering
\includegraphics[width=1.1\linewidth]{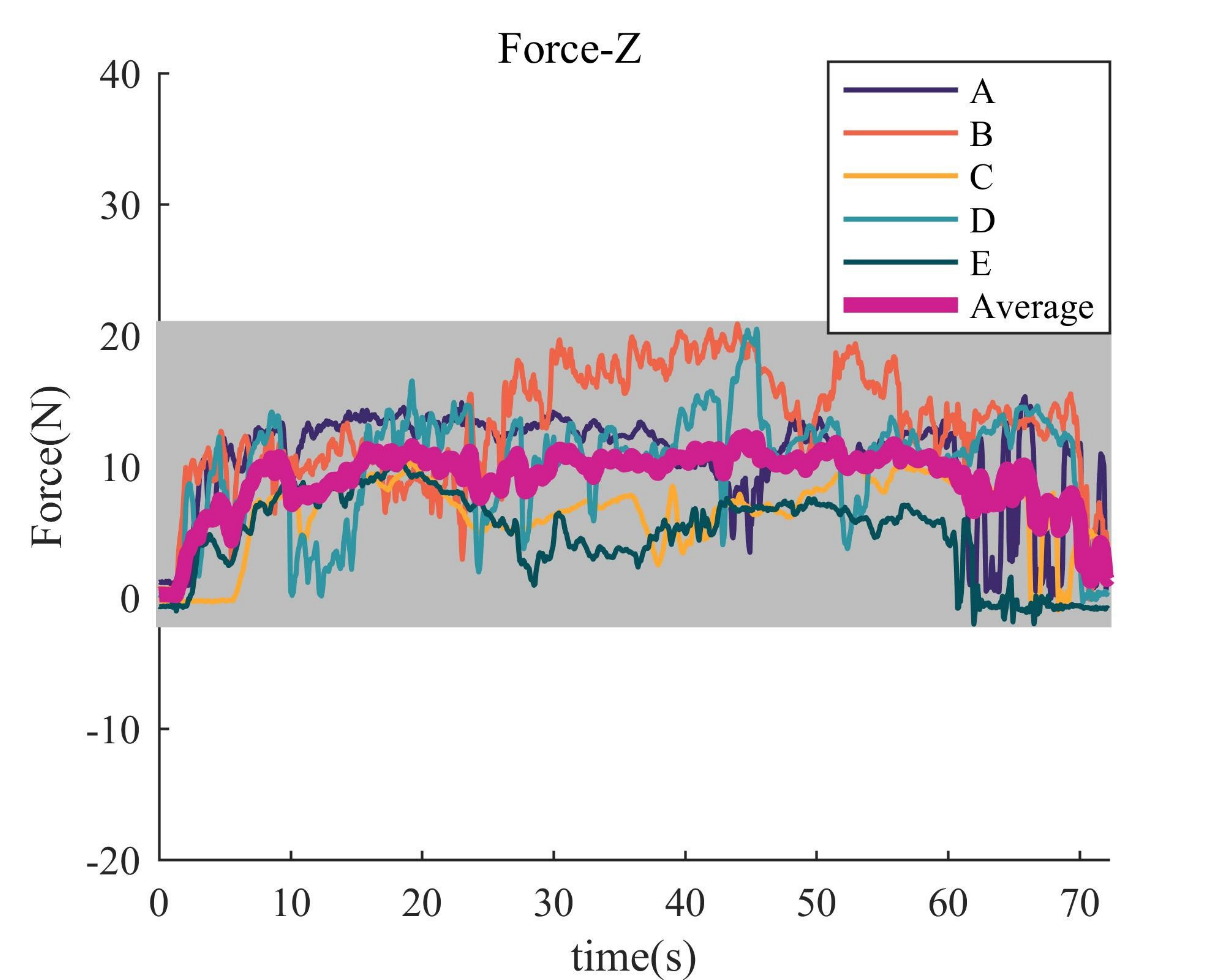}
\caption*{(c)}
\end{minipage}
\begin{minipage}[t]{0.30\linewidth}
\centering
\includegraphics[width=1.1\linewidth]{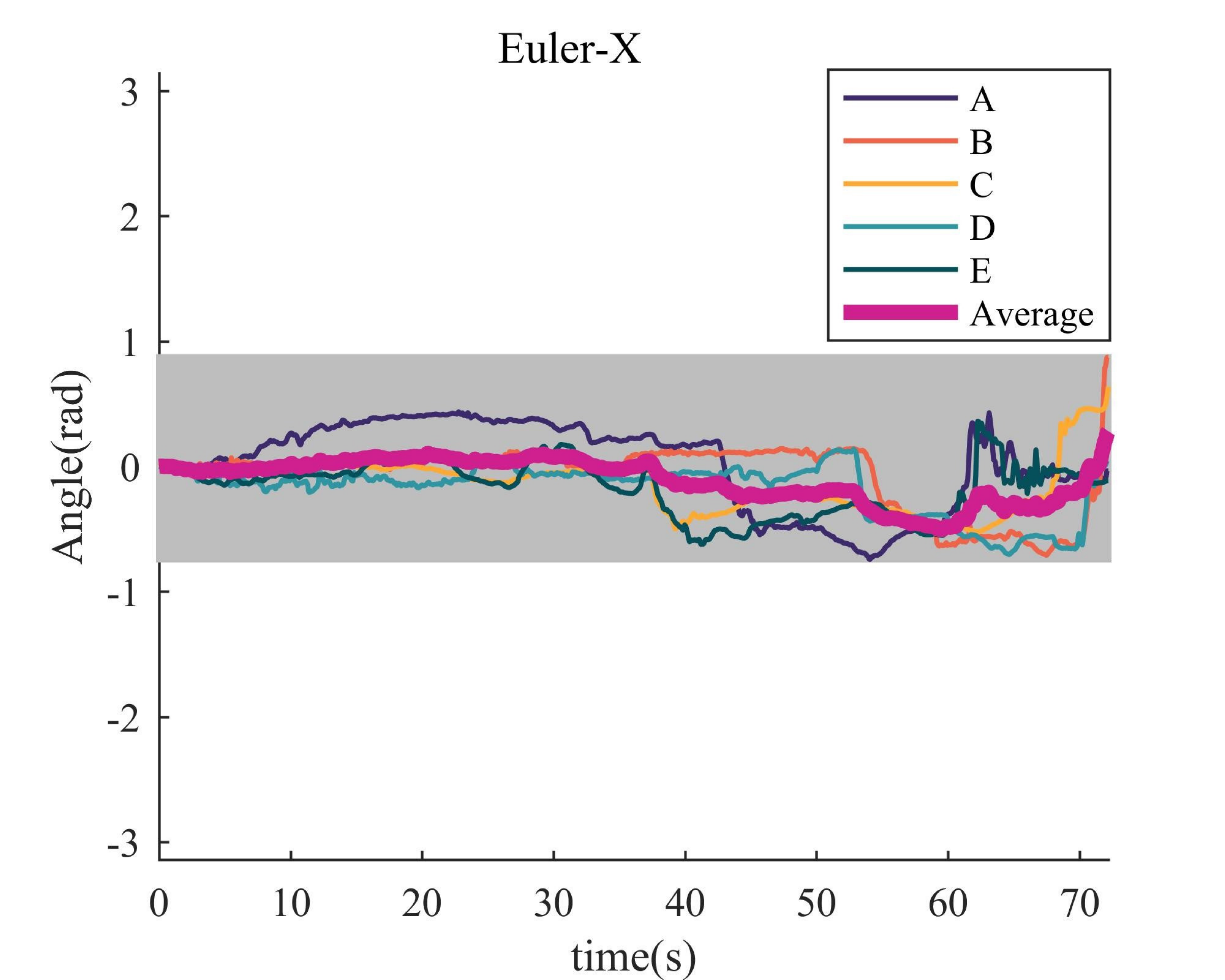}
\caption*{(d)}
\end{minipage}
\begin{minipage}[t]{0.30\linewidth}
\centering
\includegraphics[width=1.1\linewidth]{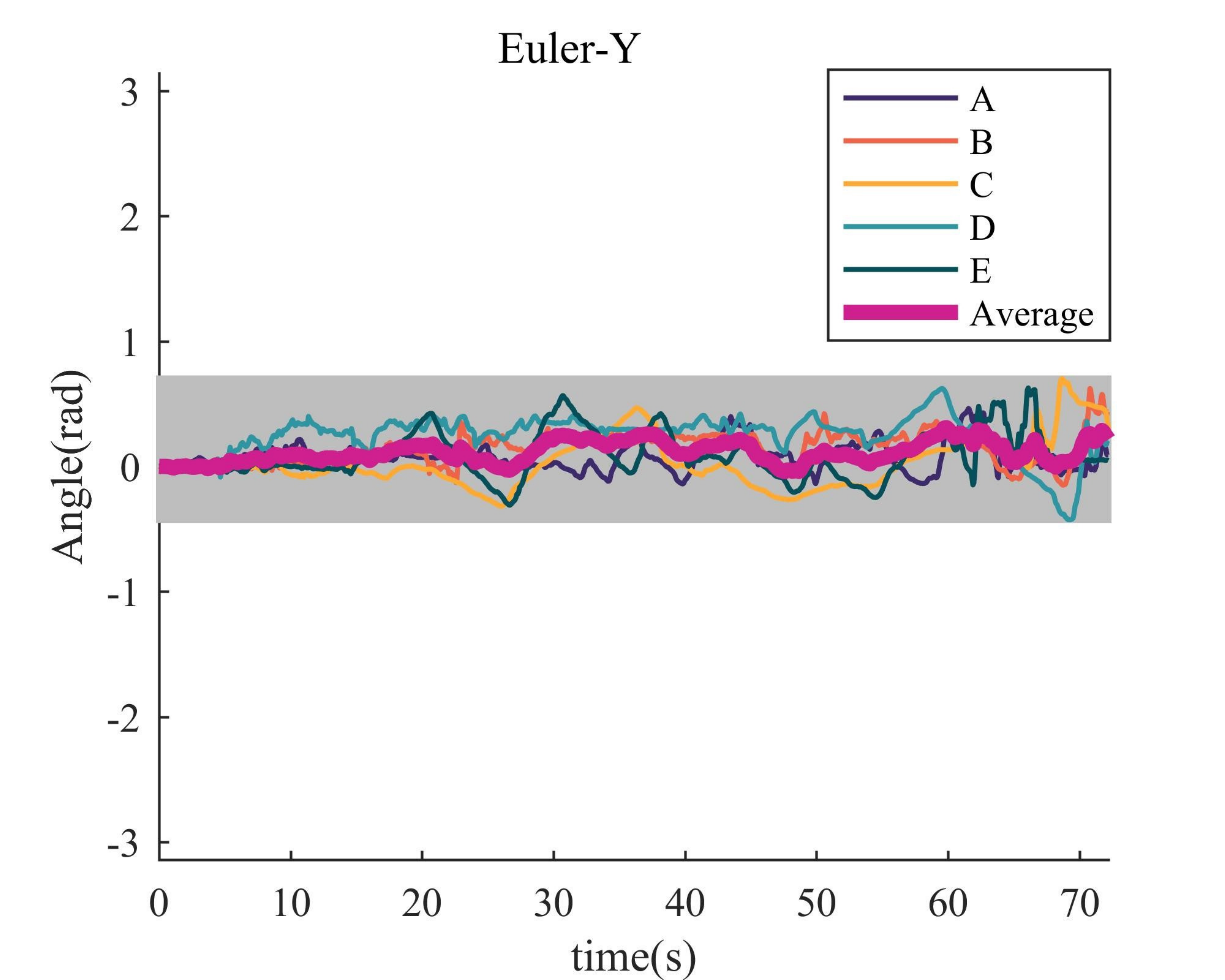}
\caption*{(e)}
\end{minipage}
\begin{minipage}[t]{0.30\linewidth}
\centering
\includegraphics[width=1.1\linewidth]{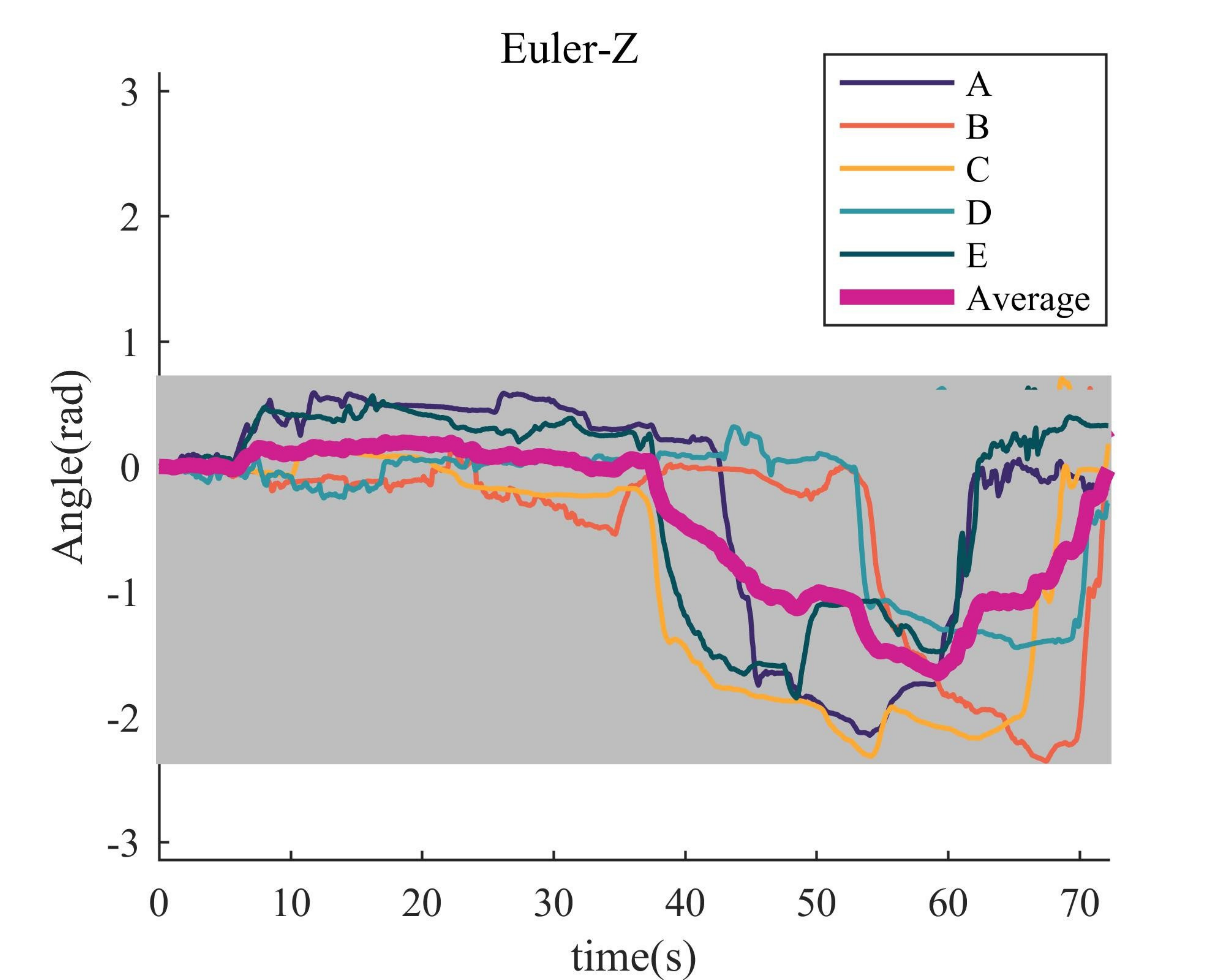}
\caption*{(f)}
\end{minipage}
\caption{The trajectories of the recorded force and pose during ultrasound examination. Force component in (a) \emph{X} direction (b) \emph{Y} direction (c) \emph{Z} direction; Rotation axis: (d) \emph{X} Axis (e) \emph{Y} Axis (f) \emph{Z} Axis.}
\label{fig::force_trj}
\end{figure*}

\subsection{Learning of Ultrasound Task Representation}
\label{sec::task_rep}
For a free-hand ultrasound scanning task, three types of sensory feedback are available: ultrasound images from the ultrasound machine, force feedback from a mounted F/T sensor, and the probe pose from a mounted IMU. To encapsulate the heterogeneous nature of this sensory data, we propose a domain-specific encoder to model the task, as shown in Fig.~\ref{fig::task_rep}. For the ultrasound imaging feedback, we use a VGG-16 network to encode the $224\times224\times3$ RGB images and yield a 128-d feature vector. For the force and pose feedback, we encode them with a 4-layer fully connected neural network to produce a 128-d feature vector. The resulting two feature vectors are concatenated together into one 256-d vector and connected with a 1-layer fully connected network to yield a 128-d feature vector as the \emph{task feature vector}. The multi-modal task representation is a neural network model denoted by $\Omega_\theta$, where the parameters are trained as described in the following section.

\subsection{Data Collection via Human Demonstration}
The multi-modal model as shown in Fig.~\ref{fig::task_rep} has a large number of learnable parameters. To obtain the training data, we design a procedure to collect the ultrasound scanning data from human demonstrations, as shown in Fig.~\ref{fig::human_demo}. A novel probe holder is designed with intrinsically mounted sensors such as IMU and F/T sensors. A sonographer is performing the ultrasound scanning process with the probe, and the data collected during the scanning process is described as follows:
\begin{itemize}
\item $D=\{(S^i,P^i,F^i)\}_{i=1...N}$ denotes a dataset with $N$ observations.
\item $S^i \in \mathbb{R}^{224\times224\times3}$ denotes the $i$-th collected ultrasound image with cropped size.
\item $P^i \in \mathbb{R}^{4}$ denotes the probe pose in terms of quaternion. 
\item $F^i \in \mathbb{R}^{6}$ denotes the $i$-th contact force/torque between the probe and the human skin.
\end{itemize}

For each recorded data in the dataset $D$, the quality of the obtained ultrasound image is evaluated by three sonographers and labeled with $1/0$. $1$ stands for a good ultrasound image while $0$ corresponds for an unacceptable ultrasound image. With the recorded data and the human annotations, the model $\Omega_\theta$ is trained with a loss function of cross-entropy. During training, we minimize the loss function with stochastic gradient descent. Once trained, this network produces a 128-d feature vector and evaluates the quality of the task at the same time. Given the task representation model $\Omega_\theta$, an online adaptation strategy is proposed to improve the task quality by leveraging the multi-modal sensory feedback, as discussed in next section.

\subsection{Ultrasound Skill Learning}
As discussed in related work, it is still challenging to model and plan complex force-relevant tasks, mainly due to the inaccurate state estimation and the lack of a dynamics model. In our case, it is difficult to explicitly model the relations among ultrasound images, the probe pose and the contact force. Therefore, we formulate the policy of ultrasound skills as a model-free reinforcement learning problem where the set of actions includes $P$ and $F$:
\begin{equation}
\begin{aligned}
& \underset{P,F}{\text{maxmize}}
& & Q_\theta=f(S,P,F|\Omega_\theta) \\
& \text{subject to}
& & P \in D_P, \; F \in D_F,\\
&&&  F_z \geq 0.
\end{aligned}
\label{eqn::formulate}
\end{equation}
where $Q_\theta$ denotes the quality of the task, which is computed using the learned model $\Omega_\theta$ by passing through the sensory feedback $S,P,F$. The constraint $F_z \geq 0$ means that the contact force along the normal direction should be positive. $D_P$ and $D_F$ denote feasible sets of the probe pose and the contact force, respectively. In our case, these two feasible sets are determined by human demonstrations. However, it is worth mentioning that other task-specific constraints for the pose and the contact force can also be adopted here.

By choosing model-free, it requires no prior knowledge of the dynamics model of the ultrasound scanning process, namely the transition probabilities from one state (current ultrasound image) to another (next ultrasound image). More specifically, we choose Monte Carlo policy optimization \cite{sutton2018reinforcement}, where the potential actions are sampled and selected directly from previous demonstrated experience, as shown in Fig.~\ref{fig::skill_sampling}. For the sampling, we impose a bound between $P^{'}_t$, $F^{'}_t$ and $P_t$, $F_t$, which prevents the next state from moving too far away from the current state. If the new state $<P^{'}_t, F^{'}_t, S_t>$ is evaluated by the task quality function $Q_\theta$ as good, thus the desired pose $P^{'}_t$ and contact force $F^{'}_t$ is used as a goal for the human ultrasound scanning guidance. Otherwise, new $P^{'}_t$ and $F^{'}_t$ are sampled from the previous demonstrated experience. This process repeat $\mathcal{N}$ times, and the $P^{'}_t$, $F^{'}_t$ with the best task quality is chosen as the final goal for the human scanning guidance. Note that this sampling-based approach does not guarantee the global optimality of Equation \ref{eqn::formulate}. However, this is sufficient for the human ultrasound scanning guidance because the final goal is only required to be updated at a relatively low frequency.

\section{Experiments: Design and Results}

\begin{figure}[ht]
\centering
\includegraphics[width=\linewidth]{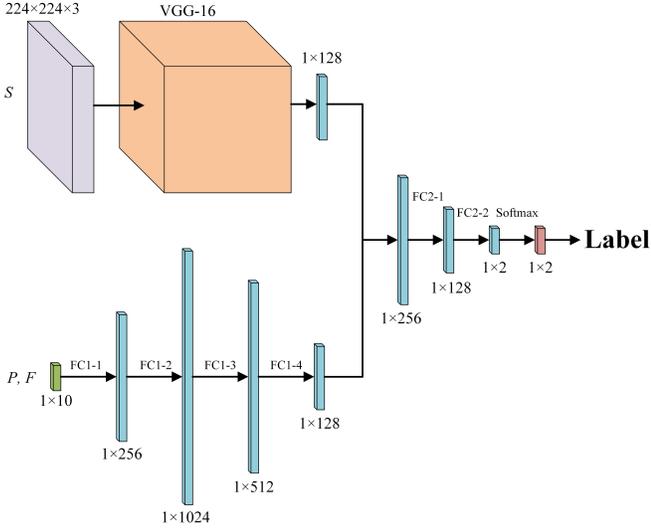}
\caption{Framework of the neural network. The ultrasound images were encoded with VGG-16. Four fully connected layers were added to transform $(P,F)$ vector into 128 channels. Vectors from $S$ and $(P,F)$ were concatenated. Two fully connected layers were added to transform concatenated vector from 256 channels to 2 channels.}
\label{fig::network}
\end{figure}
  
\begin{figure}[bp]
\centering
\includegraphics[width=\linewidth]{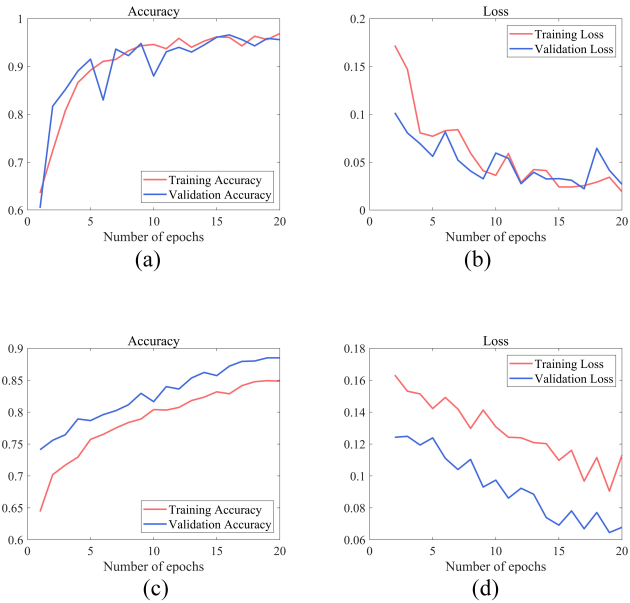}
\caption{(a) Accuracy and (b) loss in training neural network for ultrasound images classification. (c) Accuracy and (d) loss in training neural network for ultrasound skills evaluation.}
\label{fig::training}
\end{figure}
  
\begin{figure}[ht]
\centering
\includegraphics[width=\linewidth]{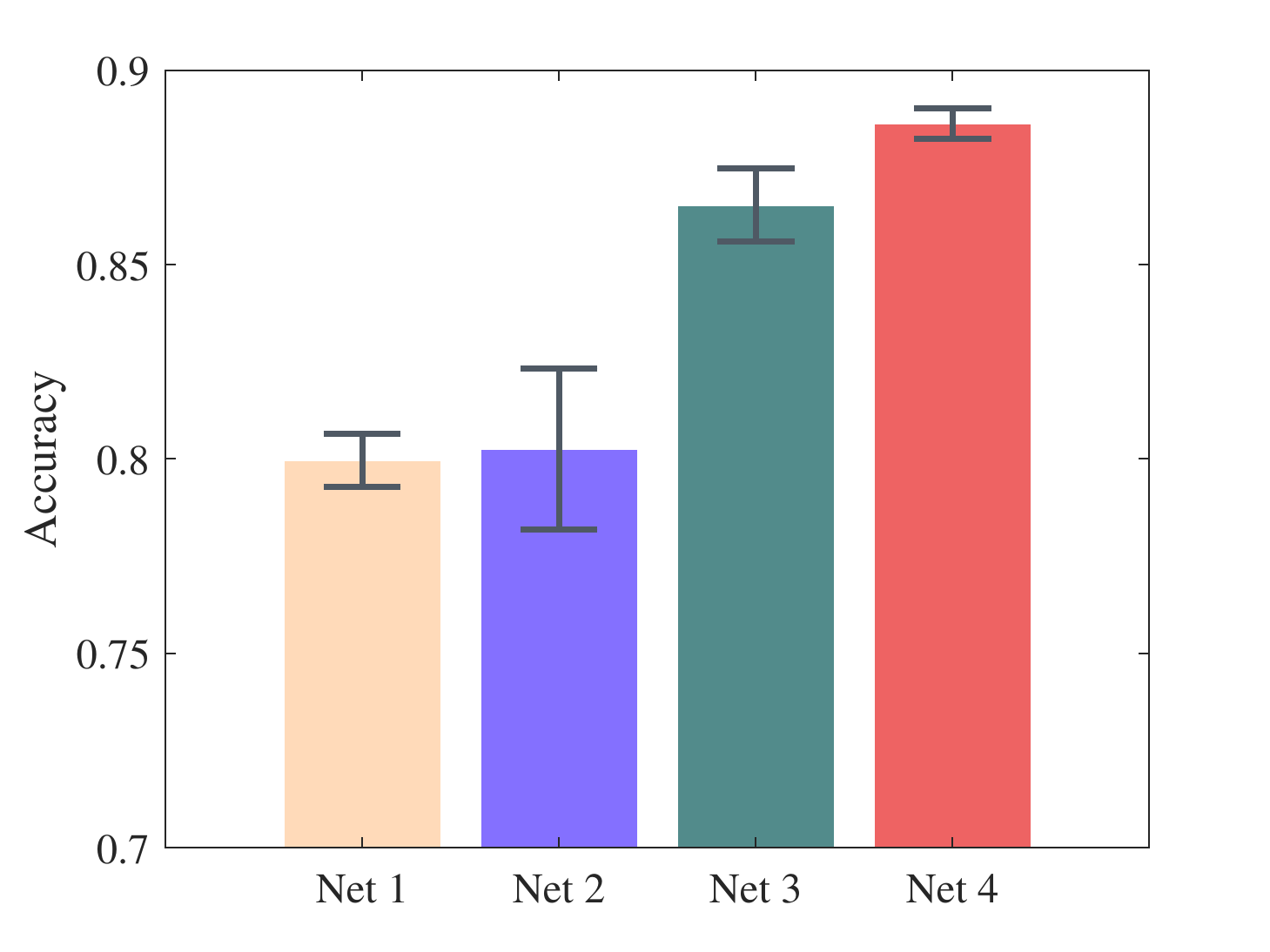}
\caption{Accuracy of four networks in validation. Net1 was trained with $S$ and $P$. Net2 was trained with $S$ and $F$. Net3 was trained with $S$, $P$ and $F$, without interaction between $P$ and $F$. Net4 was trained with $S$, $P$ and $F$, with interaction between $P$ and $F$.}
\label{fig::histogram}
\end{figure}

In this section, we use real experiments to examine the effectiveness of our proposed approach of multi-modal task representation learning. In particular, we design experiments to verify the following two questions:
\begin{itemize}
\item Does the force modality contribute to the task representation learning?
\item Is the sampling-based policy effective for real data?  
\end{itemize}

\subsection{Experiments Setup}

For the experimental setup, we used a Mindray DC-70 ultrasound machine with an imaging frame rate of 900Hz. The ultrasound image was captured using MAGEWELL USB Capture AIO with a frame rate of 120Hz and a resolution of $2048\times2160$, as shown in Fig.~\ref{fig::exp_setup}.

As shown in Fig.~\ref{fig::human_demo}, the IMU mounted on the ultrasound probe was ICM20948 and the microcontroller unit (MCU) was STM32F411. The highest frequency of IMU could reach 200Hz, with an acceleration accuracy of 0.02g and a gyroscope accuracy of $0.06^{\circ}$/s. The IMU could output the probe pose in forms of quaternion. For the force feedback, we used 6D ATI Gamma F/T sensor with a maximum frequency of 7000Hz. The computer used for the data collection was with Intel i5 CPU and Nvidia GTX 1650 GPU, and with the operation system of Ubuntu16.04 LTS and ROS Kinetic.

The ultrasound data was collected at the Hospital of Wuhan University. The sonographer was asked to scan the left kidneys of $5$ volunteers with different physical conditions. Before examination, the sonographer vertically held the probe above the left kidney of a volunteer. The ultrasound scanning process began with the recording program launched. The snapshots for the scanning process are shown in Fig.~\ref{fig::human_scan}. The collected data is consisting of ultrasound videos, the probe pose (quaternion), the contact force (force and torque) and labels (1/0). In total, there are $5995$ samples of data. The number of positive samples (labeled 1) is 2266, accounting for 37.8$\%$. The number of negative samples (labeled 0) is 3729, accounting for 62.2$\%$. Fig.~\ref{fig::force_trj} presents trajectories of the recorded information.

\begin{figure*}[ht]
\centering
\begin{minipage}[t]{0.3\textwidth}
\centering
\includegraphics[width=1\linewidth]{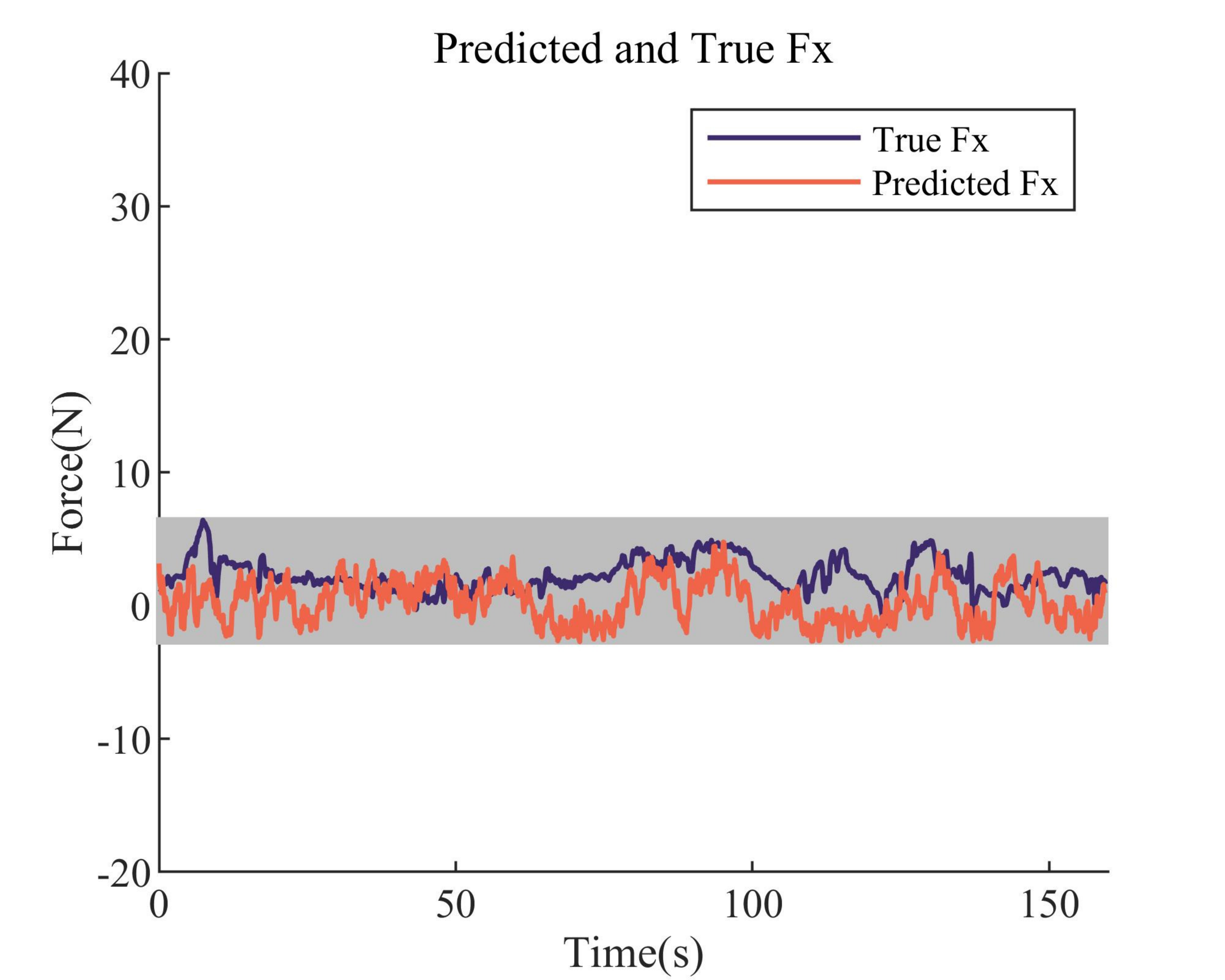}
\caption*{(a)}
\end{minipage}
\begin{minipage}[t]{0.3\textwidth}
\centering
\includegraphics[width=1\linewidth]{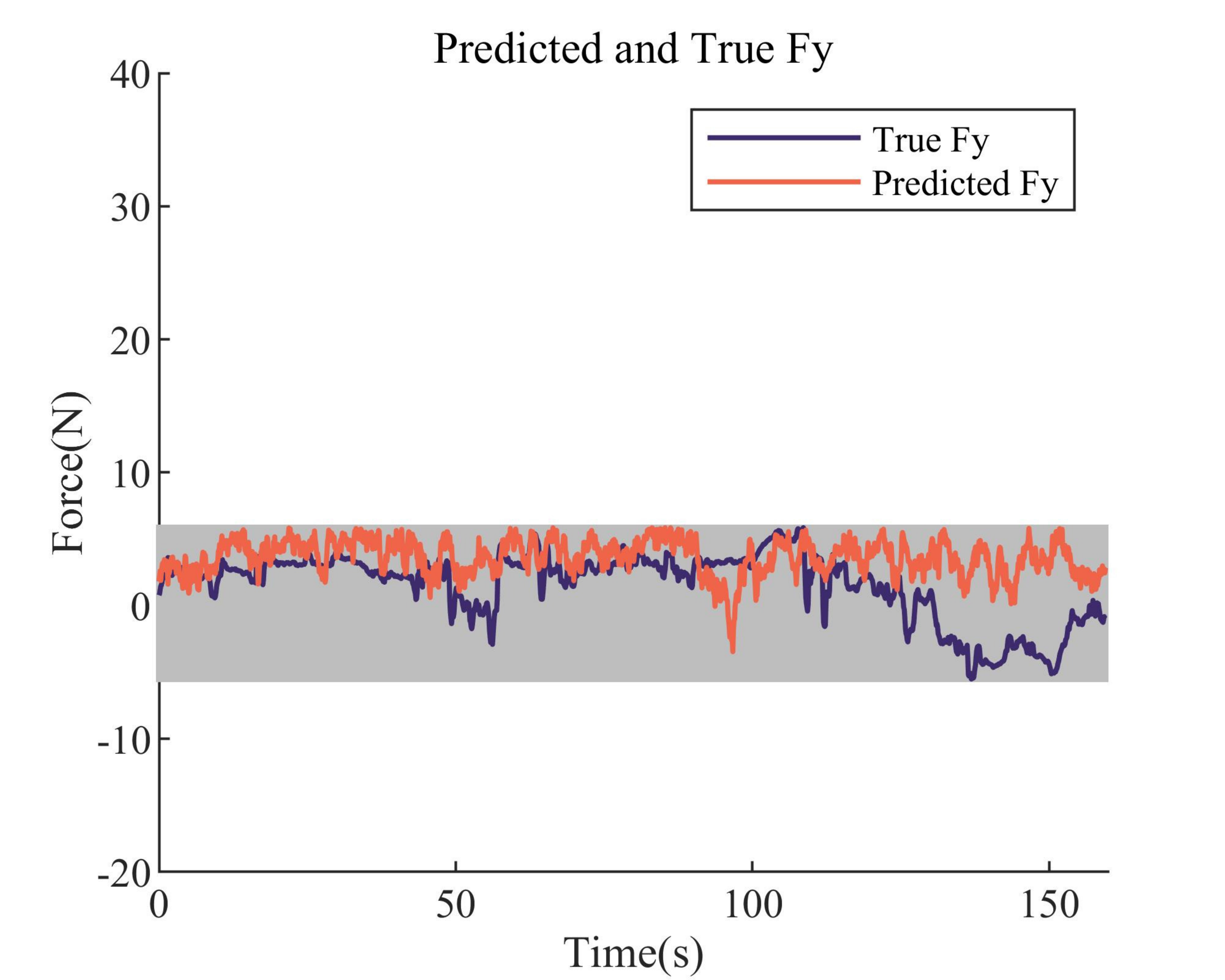}
\caption*{(b)}
\end{minipage}
\begin{minipage}[t]{0.3\textwidth}
\centering
\includegraphics[width=1\linewidth]{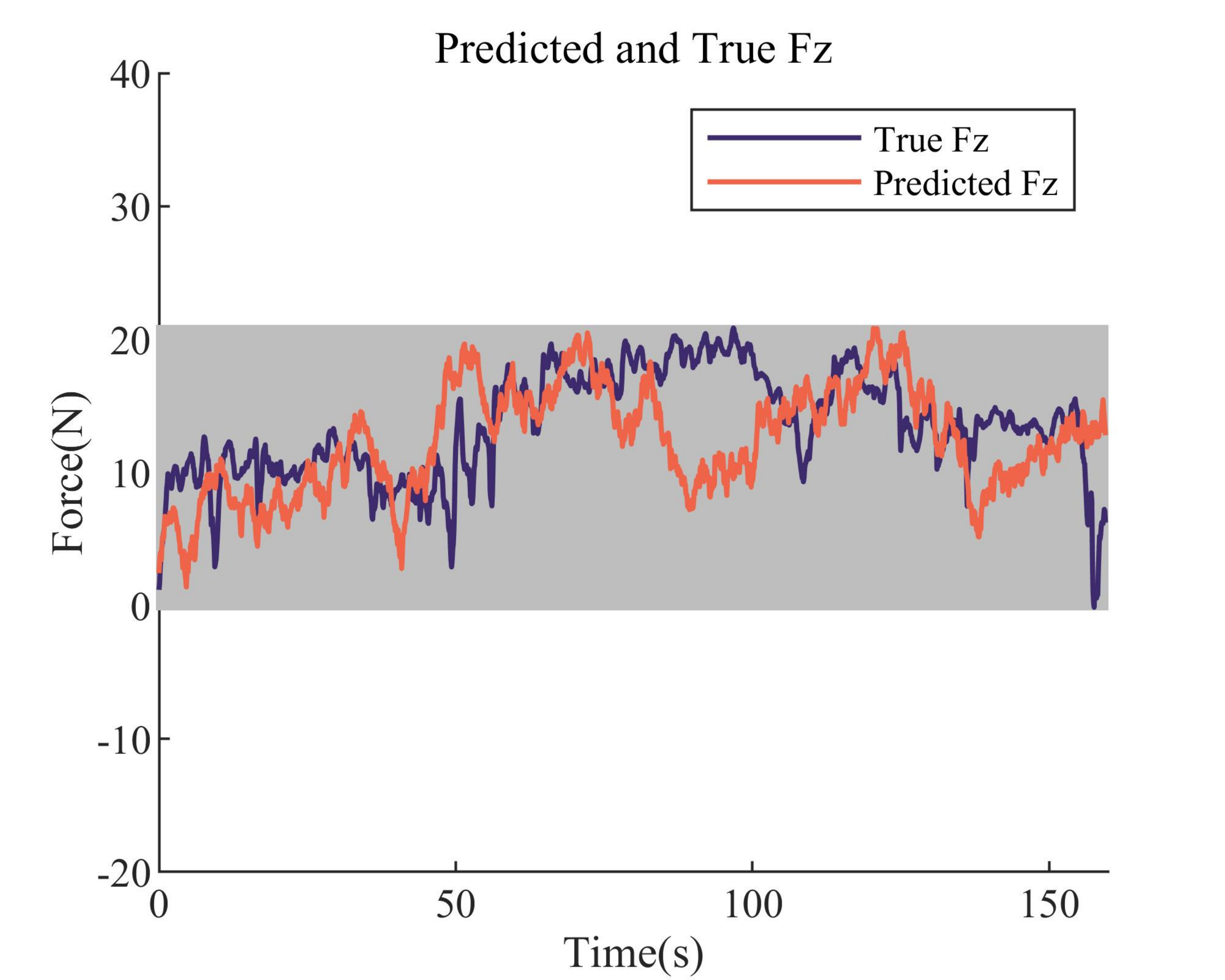}
\caption*{(c)}
\end{minipage}
\caption{Predicted force's component in (a) \emph{X} axis direction. (b) \emph{Y} axis direction. (c) \emph{Z} axis direction.}
\label{fig::pred_force}
\end{figure*}

\begin{figure*}[bp]
\centering
\includegraphics[width=1\textwidth]{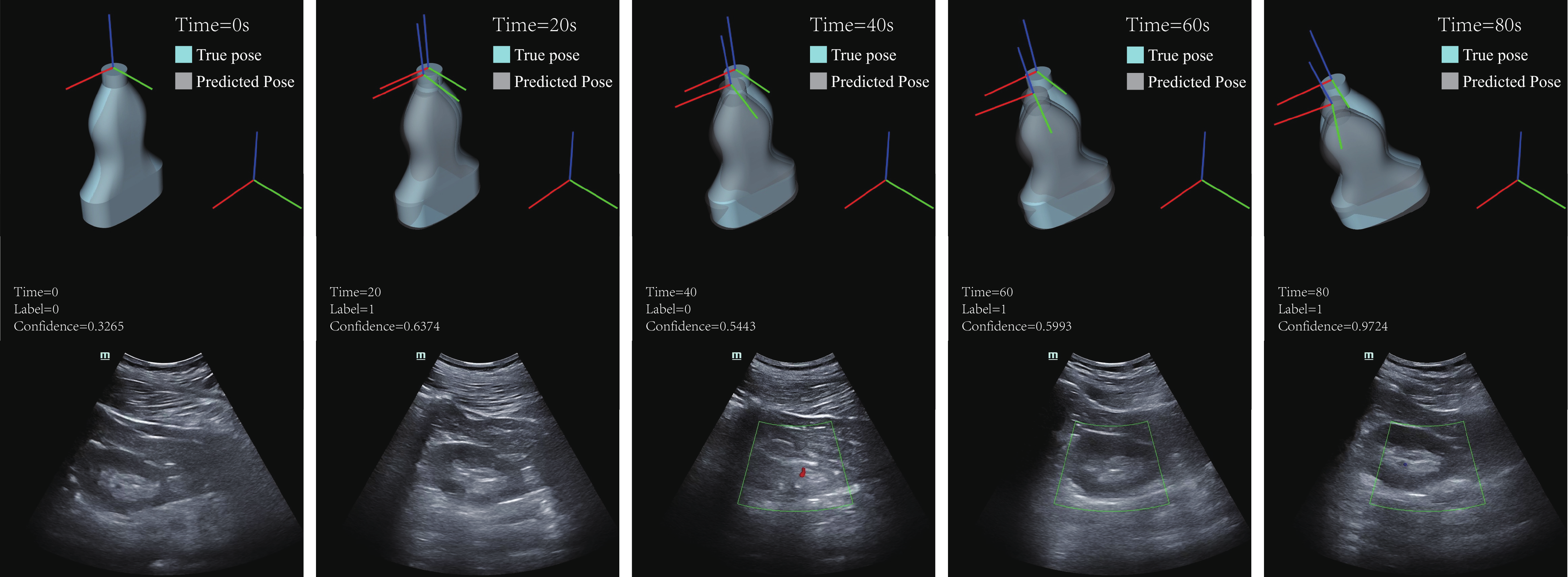}
\caption{Predicted probe pose and corresponding ultrasound images. The confidence is the probability of label 1.}
\label{fig::pred_pose}
\end{figure*}

\begin{figure*}[ht]
\centering
\includegraphics[width=1\textwidth]{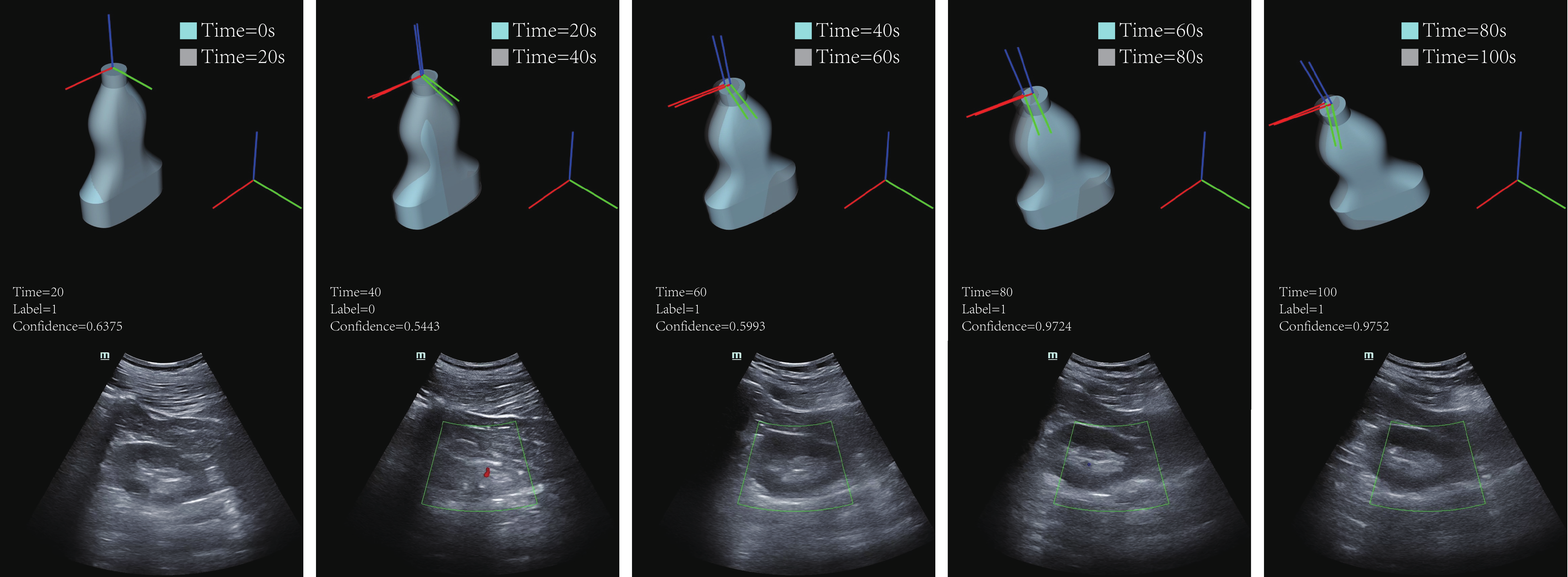}
\caption{Predicted and true probe pose, with corresponding ultrasound images. The confidence is the probability of label 1.}
\label{fig::pred_true_pose}
\end{figure*}

\subsection{Experimental Results}

The detailed architecture of our network is shown in Fig.~\ref{fig::network}. We started the training process with a warm start to classify the ultrasound images. The adopted neural network was VGG-16 with cross entropy loss. Data for training included ultrasound images and labels. The learning rate was 0.001 and the batch size was 20. For the ultrasound skill evaluation, data for training included images $S$, quaternion $P$, force $F$ and labels. By inputting $P,F,S$, this neural network would output predicted label. We fixed channels of last fully connected layer in VGG-16 to 128 channels, and merged it with $(P,F)$ feature vector. Four fully connected layers were added to transform $(P,F)$ vector into 128 channels, which concatenated with VGG-16 output vector. After getting the vector with 256 channels, two fully connected layers and a softmax layer were added to output the confidence of label. Fig.~\ref{fig::training} presents accuracy and loss in training neural networks. The neural network for classification finally reached accuracy of 96.89$\%$ and 95.61$\%$ in dataset of training and validation. The neural network for ultrasound skill evaluation finally reached accuracy of 84.85$\%$ and 88.50$\%$ in training and validation.

To confirm the correlation between $P$ and $F$, we divided data into different levels for training of four networks with different input ports. Net1 was trained with $S$ and $P$, while Net2 was trained with $S$ and $F$. Net3 was trained with $S$, $P$ and $F$ with two parallel 4-layer fully connected neural networks for inputting $P$ and $F$. Net4 (Fig.~\ref{fig::network}) was trained with $S$, $P$ and $F$, with concatenated $(P,F)$ vectors. The main difference between Net3 and Net4 was the existence of interactions between $P$ and $F$ during training process. Each network had been trained for five times with 20 training epoches. Fig.~\ref{fig::histogram} presents performance of four networks in validation.

Online ultrasound scanning skill guidance: We selected some continuous data stream from dataset for verification, which had not been used for training the neural network. The sampling process in Fig.~\ref{fig::skill_sampling} was repeated $1000$ times and the actions $P,F$ with the best task quality was selected as the next desired action. The whole process took 3 to 5 seconds to output the desired action.

Fig.~\ref{fig::pred_force} presents predicted results about components of contact force, compared with ground truth data. Fig.~\ref{fig::pred_pose} presents the predicted probe pose with corresponding ultrasound images. Fig.~\ref{fig::pred_true_pose} presents predicted and true probe pose with corresponding ultrasound images.

\section{Discussion and Conclusion}
\subsection{Discussion}
There are some limitations in this paper. First, the online guidance method is based on random sampling, which leads to a certain degree of randomness. Therefore, there is a certain difference between forecast results and true values in the short term. Second, to ensure the effectiveness of the sampling, a large number of samples are required, which means a higher task quality improvement would require more computation cost. With the expendation of dataset, this method is difficult to meet the requirement of timely guidance, which can be solved by denoting the feasible set as a probabilistic model to acquire better sampling efficiency. Finally, we believe that through detailed adjustments to the neural network, the efficiency of this model has the opportunity to be greatly improved without losing too much accuracy.

\subsection{Conclusion}
This paper presents a framework for learning ultrasound scanning skills from human demonstrations. By analyzing the scanning process of sonographers, we define the entire scanning process as a multi-modal model of interactions between ultrasound images, the probe pose and the contact force. A deep-learning based method is proposed to learn ultrasound scanning skills, from which a sampling-based strategy for ultrasound scanning guidance is proposed. Experimental results show that this framework for ultrasound scanning guidance is robust, and presents the possibility of developing a real-time learning guidance system. In the future work, we will speed up the prediction process by taking advantage of self-supervision, with the goal to port the learned guidance model into a real robot system.

\bibliographystyle{IEEEtran}
\bibliography{reference}

\end{document}